\documentclass[10pt,twocolumn,letterpaper]{article}

\usepackage[pagenumbers]{cvpr} 
\usepackage{pifont}
\usepackage{graphicx}
\usepackage{overpic}

\usepackage[dvipsnames]{xcolor}

\definecolor{cvprblue}{rgb}{0.21,0.49,0.74}
\usepackage[pagebackref,breaklinks,colorlinks,citecolor=cvprblue]{hyperref}
\usepackage{multirow}
\usepackage{booktabs}

\def\paperID{***} 
\def\confName{CVPR}
\def\confYear{2024}

\title{EditGuard: Versatile Image Watermarking for Tamper Localization and Copyright Protection
}

\author{
    Xuanyu Zhang\quad Runyi Li\quad Jiwen Yu\quad Youmin Xu\quad Weiqi Li\quad Jian Zhang$^{\dagger}$\\
    School of Electronic and Computer Engineering, Peking University, Shenzhen, China
}

\begin{document}

\twocolumn[{
\renewcommand\twocolumn[1][]{#1}
\maketitle
\centering
\vspace{-0.8cm}
\includegraphics[width=\textwidth]{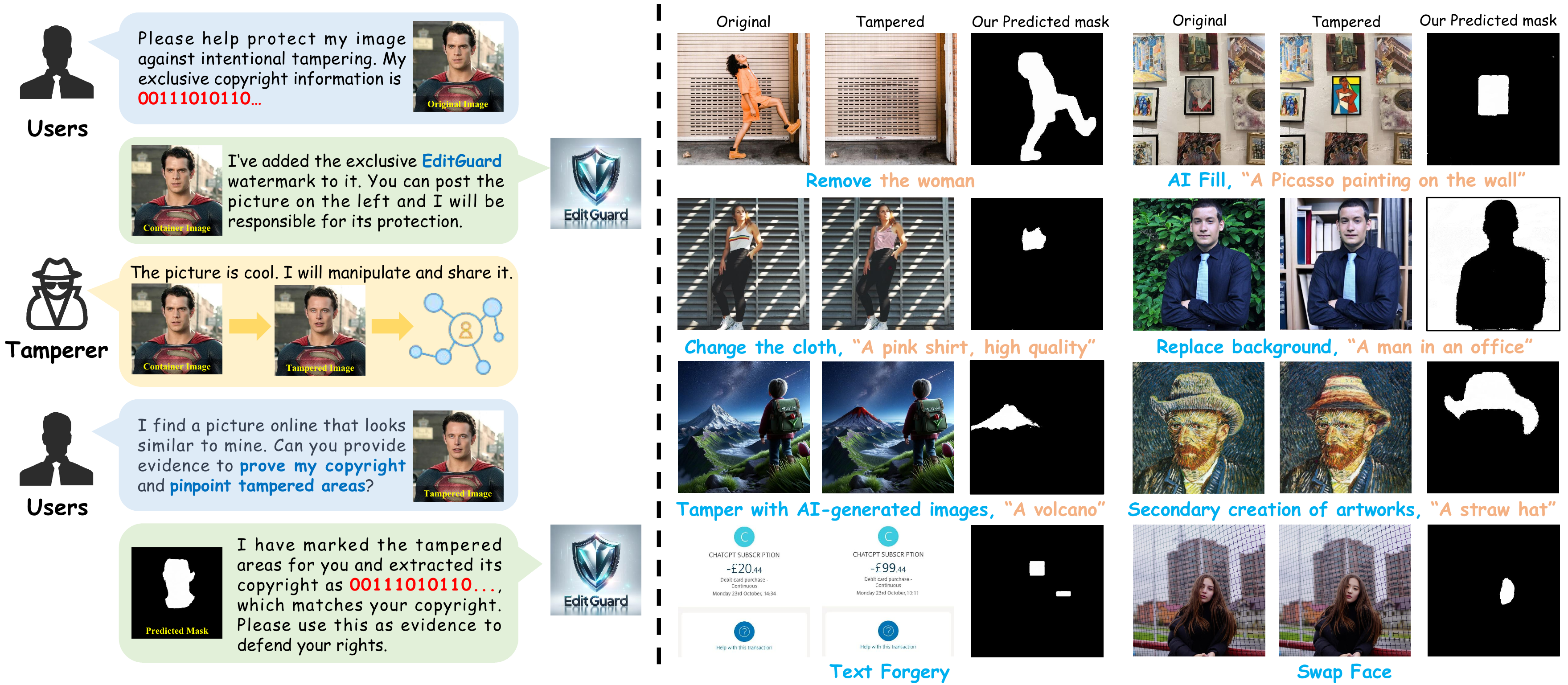}
\vspace{-0.6cm}
\captionsetup{type=figure}
\caption{We propose a versatile proactive forensics framework \textbf{EditGuard}. The application scenario is shown on the left, wherein users embed invisible watermarks to their images via EditGuard in advance. If suffering tampering, users can defend their rights via the tampered areas and copyright information provided by EditGuard. Some supported tampering methods \textcolor{blue}{(marked in blue)} and localization results of EditGuard are placed on the right. Our EditGuard can achieve over \textcolor{blue}{95\%} localization precision and nearly \textcolor{blue}{100\%} copyright accuracy.} 
}
\vspace{0.4cm}
]

\renewcommand*{\thefootnote}{$\dagger$}
\footnotetext[1]{Corresponding author.}

\begin{abstract}

In the era where AI-generated content (AIGC) models can produce stunning and lifelike images, the lingering shadow of unauthorized reproductions and malicious tampering poses imminent threats to copyright integrity and information security. Current image watermarking methods, while widely accepted for safeguarding visual content, can only protect copyright and ensure traceability. They fall short in localizing increasingly realistic image tampering, potentially leading to trust crises, privacy violations, and legal disputes. To solve this challenge, we propose an innovative proactive forensics framework \textbf{EditGuard}, to unify copyright protection and tamper-agnostic localization, especially for AIGC-based editing methods. It can offer a meticulous embedding of imperceptible watermarks and precise decoding of tampered areas and copyright information. Leveraging our observed fragility and locality of image-into-image steganography, the realization of EditGuard can be converted into a united image-bit steganography issue, thus completely decoupling the training process from the tampering types. Extensive experiments demonstrate that our EditGuard balances the tamper localization accuracy, copyright recovery precision, and generalizability to various AIGC-based tampering methods, especially for image forgery that is difficult for the naked eye to detect. The project page is available at \href{https://xuanyuzhang21.github.io/project/editguard/}{https://xuanyuzhang21.github.io/project/editguard/}.
\vspace{-0.5cm}
\end{abstract}    
\section{Introduction}
\label{sec:intro}

Owing to the advantageous properties of diffusion models and the bolstering of extensive datasets, AI-generated content (AIGC) models like DALL·E 3~\cite{dalle3}, Imagen~\cite{saharia2022photorealistic}, and Stable Diffusion~\cite{rombach2022high}, can produce lifelike and wondrous images, which greatly facilitate the endeavors of photographers and image editors. Nonetheless, the remarkable capabilities of these models come with a double-edged sword, presenting new challenges in copyright protection and information security. The efficiency of image manipulation~\cite{rombach2022high, podell2023sdxl, zhang2023adding, lugmayr2022repaint, faceswap, yu2023freedom, suvorov2022resolution, yang2023difflle, cheng2023null, cheng2023progressive3d} has blurred the line between fact and forgery, ushering in myriad security and legal concerns. For instance, artistic works are vulnerable to malicious tampering or unauthorized AI-facilitated recreations, making it challenging to protect their original creations~\cite{copyright, battle}. Meanwhile, forged images may be spread online or used as court evidence, causing adverse effects on public opinion, ethical issues, and social stability~\cite{forged}.

Given the challenges of preventing image tampering from the source, image watermarking has become a consensus for proactive forensics~\cite{whitehouse, synthid}. However, prevalent forensic image watermarking~\cite{fernandez2023stable, wu2023sepmark, ma2022towards, synthid} still focus on detecting image authenticity or protecting image copyrights, but fall short when it comes to advanced demands, such as localizing tampered areas. Tamper localization facilitates an evaluation of the severity of the image manipulation, and provides an understanding of the intent of attackers, potentially allowing for the partial reuse of the tampered images. However, passive forensics methods such as previous black-box localization networks~\cite{dong2022mvss,sun2023safl,wu2022robust} tend to seek anomalies like artifacts or flickers in images but struggle to detect more realistic textures and more advanced AIGC models. Moreover, they inevitably need to introduce tampered data during the training and focus solely on specific ``CheapFake'' tampering like slicing and copy-and-paste~\cite{dong2022mvss, sun2023safl, ma2023iml}, or on ``DeepFake'' targeting human faces~\cite{ramesh2022hierarchical, asnani2023malp}, restricted in generalizability. Thus, it is vital to develop an integrated watermarking framework that unites tamper-agnostic localization and copyright protection.

To clarify our task scope, we re-emphasize the definition of dual forensics tasks as illustrated in Fig.~\textcolor{red}{1}: \textbf{\underline{(1) Copyright protection:}} ``Who does this image belong to?'' We aspire to accurately retrieve the original copyright of an image, even suffering various tampering and degradation. \textbf{\underline{(2) Tamper localization}:} ``Where was this image manipulated?'' We aim to precisely pinpoint the tampered areas, unrestrained by specific tampering types. To the best of our knowledge, no existing method accomplishes these two tasks simultaneously, while maintaining a balance of high precision and extensive generalizability.


To address this urgent demand, we propose a novel proactive forensics framework, dubbed \textbf{EditGuard}, to protect copyrights and localize tamper areas for AIGC-based editing methods. Specifically, drawing inspiration from our observed locality and fragility of image-into-image (I2I) steganography and inherent robustness of bit-into-image steganography, we can transform the realization of EditGuard into a joint image-bit steganography issue, which allows the training of EditGuard to be entirely decoupled from tampering types, thereby endowing it with exceptional generalizability and locate tampering in a zero-shot manner. In a nutshell, our contributions are as follows:

\vspace{2pt}
\noindent \ding{113}~(1) We present the first attempt to design a versatile proactive forensics framework \textbf{EditGuard} for universal tamper localization and copyright protection. It embeds dual invisible watermarks into original images and accurately decodes tampered areas and copyright information. 

\vspace{2pt}
\noindent \ding{113}~(2) We have observed the fragility and locality of I2I steganography and innovatively convert the solution of this dual forensics task into training a united Image-Bit Steganography Network (IBSN), and utilize the core components of IBSN to construct EditGuard.

\vspace{2pt}
\noindent \ding{113}~(3) We introduce a prompt-based posterior estimation module to enhance the localization accuracy and degradation robustness of the proposed framework.

\vspace{2pt}
\noindent \ding{113}~(4) The effectiveness of our method has been verified on our constructed dataset and classical benchmarks. Compared to other competitive methods, our approach has notable merits in localization precision, generalization abilities, and copyright accuracy without any labeled data or additional training required for specific tampering types.

\section{Related works}
\label{sec:related}
\subsection{Tamper Localization}
Prevalent passive image forensic techniques have focused on localizing specific types of manipulations~\cite{wu2022robust, salloum2018image, islam2020doa, li2018fast, zhu2018deep, li2019localization}. Meanwhile, some universal tamper localization methods~\cite{li2018learning, kwon2021cat, chen2021image, wu2019mantra, ying2023learning, ying2021image, hu2023draw, ying2022rwn} also tend to explore artifacts and anomalies in tampered images. For instance, MVSS-Net~\cite{dong2022mvss} employed multi-view feature learning and multi-scale supervision to jointly exploit boundary artifacts and the noise view of images. OSN~\cite{wu2022robust} proposed a novel robust training scheme to address the challenges posed by lossy operations. Trufor~\cite{guillaro2023trufor} used a learned noise-sensitive fingerprint and extracted both high-level and low-level traces via transformer-based fusion. HiFi-Net~\cite{guo2023hierarchical} utilized multi-branch feature extractor and localization modules for both CNN-synthesized and edited images. SAFL-Net~\cite{sun2023safl} constrained a feature extractor to learn semantic-agnostic features with specific modules and auxiliary tasks. However, the above-mentioned passive localization methods are often limited in terms of generalization and localization accuracy, which usually work on known tampering types that have been trained. Although MaLP~\cite{asnani2023malp} used template matching for proactive tamper localization, it still requires a large number of forgery images and cannot fully decouple the network training from the tamper types.
\begin{figure}[t!]
	\centering
	\includegraphics[width=1\linewidth]{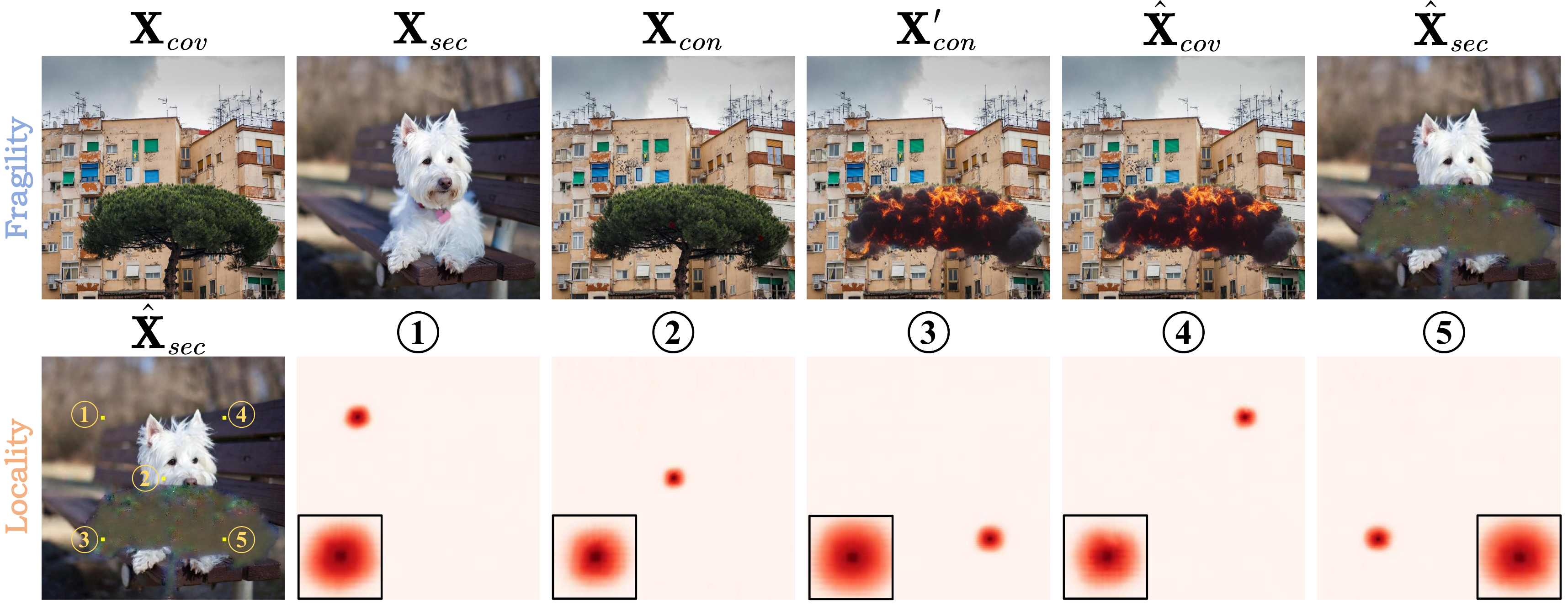}
	\vspace{-15pt}
	\caption{Fragility and locality of I2I steganography. 
The first line shows that when $\mathbf{X}^{\prime}_{con}$ is changed, $\hat{\mathbf{X}}_{sec}$ will also be fragilely demaged. The second line plots the attribution maps $\frac{1}{|\mathbb{S}|}\sum_{(i, j) \in \mathbb{S}} \frac{\partial\hat{\mathbf{X}}_{sec}[i, j]}{\partial\mathbf{X}^{\prime}_{con}}$ of five point sets $\mathbb{S}$ (marked by \textcircled{1}-\textcircled{5}) in $\hat{\mathbf{X}}_{sec}$. We observed that $\hat{\mathbf{X}}_{sec}$ almost only has a strong response at the corresponding positions of $\mathbf{X}^{\prime}_{con}$ and its neighborhoods.}
    \vspace{-15pt}
    \label{local}
\end{figure}
\subsection{Image Watermarking}
Image watermarking~\cite{wang2023security} can be broadly divided into two types based on their purposes, namely adversarial watermarking and forensic watermarking. Adversarial watermarking~\cite{liang2023adversarial, van2023anti, ye2023duaw, ma2023generative, feng2023catch} intentionally confuse generative models via embedding perturbations into images, thus creating anomalous adversarial examples. Forensic watermarking~\cite{zhu2018hidden}, on the other hand, is used for the verification, authenticity, and traceability of images. Previous methods tend to use deep encoder-decoder networks~\cite{zhu2018hidden, liu2019novel, ahmadi2020redmark, neekhara2022facesigns, wu2023sepmark} or flow-based models~\cite{fang2023flow, ma2022towards} to hide and recover bitstream. Recently, researchers~\cite{wen2023tree,fernandez2023stable,jiang2023evading, zhao2023recipe, cui2023diffusionshield} have designed specialized watermarking mechanisms for large-scale image generation models, such as stable diffusion~\cite{rombach2022high}, to merge watermarking into the generation process. However, these watermarking methods have a singular function and cannot accurately localize the tampered areas.

\begin{figure*}[t!]
	\centering
	\includegraphics[width=1\linewidth]{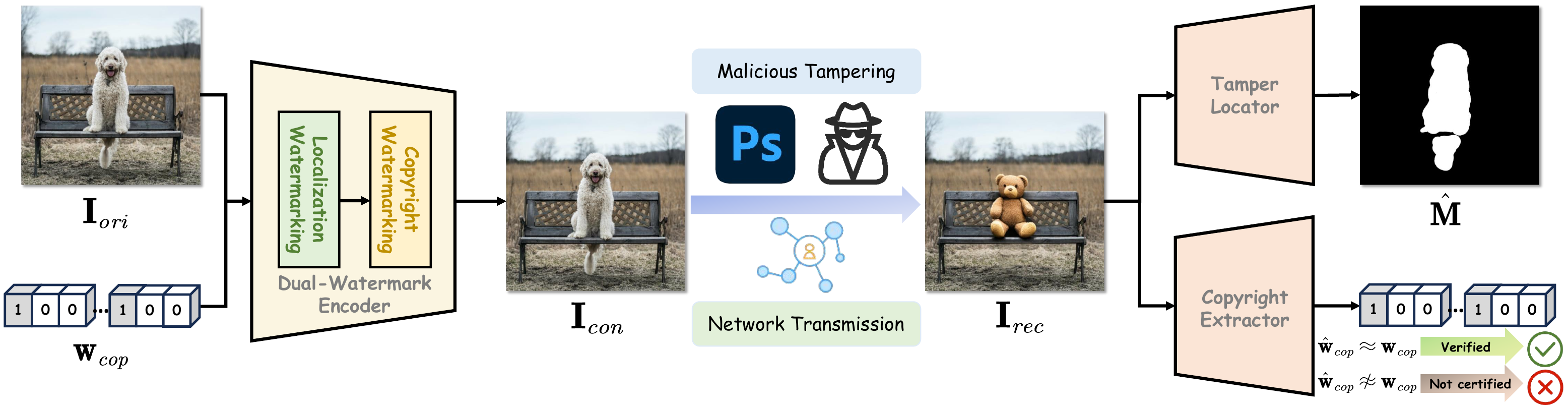}
	\vspace{-15pt}
	\caption{Illustration of the proposed proactive forensics framework EditGuard. The dual-watermark encoder sequentially embeds the pre-defined localization watermark and copyright watermark $\mathbf{w}_{cop}$ into the original image $\mathbf{I}_{ori}$, generating the container image $\mathbf{I}_{con}$. After encountering potential malicious tampering and degradation during network transmission, tampered mask $\hat{\mathbf{M}}$ and copyright information $\hat{\mathbf{w}}_{cop}$ are respectively extracted via the tamper locator and copyright extractor from the received image $\mathbf{I}_{rec}$.}
	\label{editguard}
     \vspace{-15pt}
\end{figure*}


\section{Overall Framework of EditGuard}
\label{sec:problem}
\subsection{Motivation}
\label{motivation}
\textbf{Challenges of existing methods:} (1) How to equip existing watermark methods, which are solely for copyright protection, with the ability to localize tampering is the crux of EditGuard. We will solve it via the framework design in Sec.~\ref{frameworkdesign}. (2) Most previous tamper localization methods inevitably introduce specific tampering data during network training but tend to raise generalization concerns in unknown tampering types, which will be addressed in Sec.~\ref{transform}.


\noindent \textbf{Our observations:} Fortunately, we observed that image-into-image (I2I) steganography exhibits distinct fragility and locality, possessing great potential to address these issues. Concretly, I2I steganography~\cite{baluja2017hiding, xu2022robust, mou2023large, lu2021large, jing2021hinet, yu2023cross} aims to hide a secret image $\mathbf{X}_{sec}$ into a cover image $\mathbf{X}_{cov}$ to produce a container image $\mathbf{X}_{con}$, and reveal $\hat{\mathbf{X}}_{sec}$ and $\hat{\mathbf{X}}_{cov}$ with minimal distortion from the received image $\mathbf{X}^{\prime}_{con}$. We discover that when $\mathbf{X}^{\prime}_{con}$ undergoes significant alteration compared to $\mathbf{X}_{con}$, $\hat{\mathbf{X}}_{sec}$ will also be damaged and generate artifacts (the first row of Fig.~\ref{local}), which is called \textcolor{blue}{\textbf{\textit{fragility}}}. Furthermore, we notice that the artifacts in $\hat{\mathbf{X}}_{sec}$ are almost pixel-level corresponding to the changes in $\mathbf{X}^{\prime}_{con}$ relative to $\mathbf{X}_{con}$, which is called \textcolor{blue}{\textbf{\textit{locality}}}. To demonstrate this locality, we select five $7$$\times$$7$ point sets on $\hat{\mathbf{X}}_{sec}$ and calculated their attribution maps with respect to $\mathbf{X}^{\prime}_{con}$. As plotted in the second row of Fig.~\ref{local}, $\hat{\mathbf{X}}_{sec}$ only exhibits a strong response at the corresponding locations of $\mathbf{X}^{\prime}_{con}$ and their immediate vicinity, almost irrelevant to other pixels. These properties inspire us to treat $\mathbf{X}_{sec}$ as a special localization watermark and embed it within existing watermarking frameworks. 

\subsection{Framework Design and Forensics Process}
\label{frameworkdesign}
To realize united tamper localization and copyright protection, EditGuard is envisioned to embed both a 2D localization watermark and a 1D copyright traceability watermark into the original image in an imperceptible manner, which allows the decoding end to obtain the copyright of the images and a binary mask reflecting tampered areas. However, designing such a framework needs to solve the compatibility issue of two types of watermarks.

\textbf{(1) Local vs. Global}: The localization watermark is required to be hidden in the corresponding pixel positions of the original image, while the copyright watermark should be unrelated to spatial location and embedded in the global area redundantly. \textbf{(2) Semi-fragile vs. Robust}: The desired attribute of the localization watermark is semi-fragile, which means it is fragile to tampering but robust against some common degradations (such as Gaussian noise, JPEG compression, and Poisson noise) during network transmission. However, the copyright should be extracted nearly losslessly, irrespective of tampering or degradation. 

To address the two pivotal conflicts, EditGuard employs a \textbf{\textcolor{blue}{``sequential encoding and parallel decoding''}} structure, which comprises a dual-watermark encoder, a tamper locator, and a copyright extractor. As shown in Fig.~\ref{editguard}, the dual-watermark encoder will sequentially add pre-defined localization watermark and global copyright watermark $\mathbf{w}_{cop}$ provided by users to the original image $\mathbf{I}_{ori}$, forming the container image $\mathbf{I}_{con}$. Our experiments have proved that parallel encoding cannot effectively add dual watermarks into images (in supplementary materials \textbf{\textcolor{blue}{(S.M.)}}). In contrast, sequential embedding effectively prevents cross-interference by hiding these two watermarks. Furthermore, we model the network transmission process in which the received (tampered) image $\mathbf{I}_{rec}$ is transformed from $\mathbf{I}_{con}$ as:
{\setlength\abovedisplayskip{0.2cm}
\setlength\belowdisplayskip{0.2cm}
\begin{equation}
    \mathbf{I}_{rec}=\mathcal{D}(\mathbf{I}_{con} \odot (\mathbf{1} - \mathbf{M}) + \mathcal{T}(\mathbf{I}_{con}) \odot \mathbf{M}), 
\end{equation}}where $\mathcal{T}(\cdot)$, $\mathcal{D}(\cdot)$ and $\mathbf{M}$ respectively denote the tamper function, degradation operation, and tempered mask. Moreover, the parallel decoding processes allow us to flexibly train each branch under different levels of robustness and obtain the predicted mask $\hat{\mathbf{M}}$ via the tamper locator and traceability watermark $\hat{\mathbf{w}}_{cop}$ via the copyright extractor. We can categorize the dual forensic process of EditGuard into the following scenarios.

\vspace{1.5pt}
\noindent \ding{113}~\textbf{Case 1:} If $\hat{\mathbf{w}}_{cop}$ $\not \approx$ $\mathbf{w}_{cop}$, suspicious $\mathbf{I}_{rec}$ is either not registered in our EditGuard or underwent extremely severe global tampering, rendering it unreliable as evidence.

\vspace{1pt}
\noindent \ding{113}~\textbf{Case 2:} If $\hat{\mathbf{w}}_{cop}$ $ \approx $ $ \mathbf{w}_{cop}$ and $\hat{\mathbf{M}}$ $\not \approx$ $\mathbf{0}$, suspicious $\mathbf{I}_{rec}$ has undergone tampering, disqualifying it as valid evidence. Users may infer the intention of tamperers based on $\hat{\mathbf{M}}$ and decide whether to reuse parts of the image.

\vspace{1pt}
\noindent \ding{113}~\textbf{Case 3:} If $\hat{\mathbf{w}}_{cop}$ $\approx$ $\mathbf{w}_{cop}$ and $\hat{\mathbf{M}}$ $\approx$ $\mathbf{0}$, $\mathbf{I}_{rec}$ remains untampered and trustworthy under the shield of EditGuard.
\begin{figure*}[t]
	\centering
	\includegraphics[width=1\linewidth]{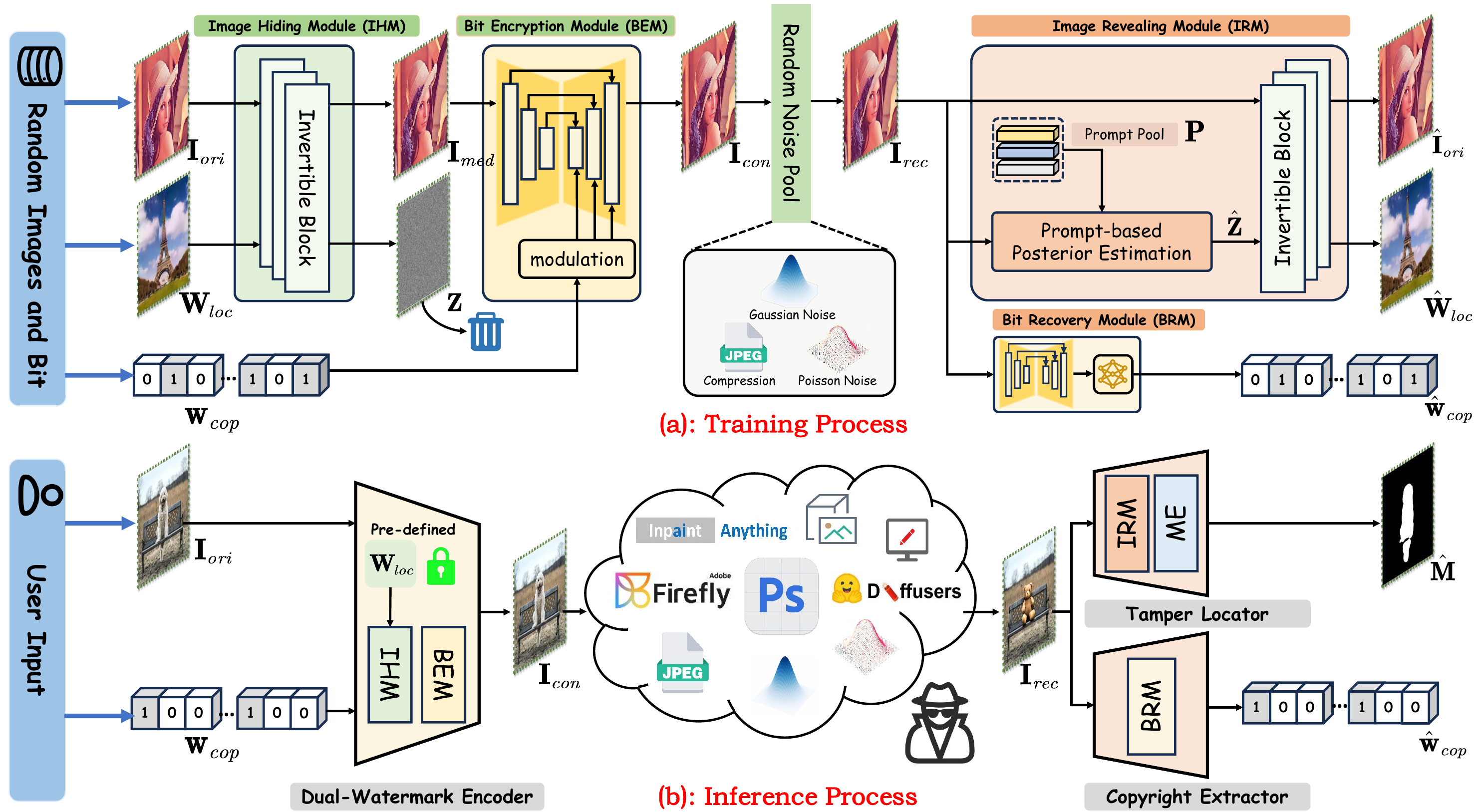}
	\vspace{-18pt}
	\caption{Illustration of the united Image-bit Steganography Network (IBSN). In the training process, we randomly sample original image $\mathbf{I}_{ori}$, localization watermark $\mathbf{W}_{loc}$ (a natural RGB image) and copyright watermark $\mathbf{w}_{cop}$ and expect the IBSN to recover $\hat{\mathbf{I}}_{ori}$, $\hat{\mathbf{W}}_{loc}$ and $\hat{\mathbf{w}}_{cop}$ with high fidelity. In the inference process, we use core components of the pre-trained IBSN with a mask extractor (ME) to construct our EditGuard, and pre-define a simple solid color image as a localization watermark.}
	\label{framework}
    \vspace{-15pt}
\end{figure*}
\subsection{Transform Dual Forensics into Steganography}
\label{transform}
To realize universal and tamper-agnostic localization, we resort to our observed locality and fragility of I2I steganography. As described in Sec.~\ref{motivation}, localization watermarking and tamper locator in Fig.~\ref{editguard} can be effectively realized via image hiding and revealing. Meanwhile, combing with the robustness of current bit-into-image steganography, copyright watermarking and extractor in Fig.~\ref{editguard} are achieved via bit encryption and recovery. Thus, we can convert the realization of the dual forensics framework EditGuard into a united image-bit steganography network. \textbf{Our training objective is just a self-recovery mechanism, meaning it only needs to ensure the input and output of the steganography network maintain high fidelity under various robustness levels, with no need to introduce any labeled data or tampered samples.} During inference, it can naturally locate tampering via simple comparisons \textbf{in a zero-shot manner} and extract copyright exactly.

\section{United Image-bit Steganography Network}
\subsection{Network Architecture}
As plotted in Fig.~\ref{framework}, the proposed IBSN includes an image hiding module (IHM), a bit encryption module (BEM), a bit recovery module (BRM), and an image revealing module (IRM). First, the IHM aims to hide a localization watermark $\mathbf{W}_{loc}$$\in$$\mathbb{R}^{H \times W \times 3}$ into the original image $\mathbf{I}_{ori}$$\in$$\mathbb{R}^{H \times W \times 3}$, resulting in an intermediate output $\mathbf{I}_{med}$$\in$$\mathbb{R}^{H \times W \times 3}$. Subsequently, $\mathbf{I}_{med}$ is fed to the BEM for feature refinement, while the copyright watermark $\mathbf{w}_{cop}$$\in$$\{0, 1\}^L$ is modulated into the BEM, forming the final container image $\mathbf{I}_{con}$$\in$$\mathbb{R}^{H \times W \times 3}$. After network transmission, the BRM will faithfully reconstruct the copyright watermark $\hat{\mathbf{w}}_{cop}$ from the received container image $\mathbf{I}_{rec}$. Meanwhile, $\mathbf{I}_{rec}$ predicts the missing information $\hat{\mathbf{Z}}$ via the prompt-based posterior estimation and uses it as the initialization for the invertible blocks, producing $\hat{\mathbf{I}}_{ori}$ and semi-fragile watermark $\hat{\mathbf{W}}_{loc}$.

\subsection{Invertible Blocks in IHM and IRM} 
Given the inherent capacity of flow-based models to precisely recover multimedia information, we harness stacked invertible blocks to construct image hiding and revealing modules. The original image $\mathbf{I}_{ori}$$\in$$\mathbb{R}^{H \times W\times 3}$ and localization watermark $\mathbf{W}_{loc}$$\in$$\mathbb{R}^{H \times W \times 3}$ will undergo discrete wavelet transformations (DWT) to yield frequency-decoupled image features. We then employ enhanced additive affine coupling layers to project the original image and its corresponding localization watermark branches. The transformation parameters are generated from each other. The enhanced affine coupling layer is composed of a five-layer dense convolution block~\cite{mou2023large} and a lightweight feature interaction module (LFIM)~\cite{chen2022simple}. The LFIM can enhance the non-linearity of transformations and capture the long-range dependencies with low computational cost. More details are presented in \textbf{\textcolor{blue}{S.M.}}. Finally, the revealed features are then transformed to the image domain via the inverse DWT. 

\subsection{Prompt-based Posterior Estimation}

\begin{figure}[t]
	\centering
	\includegraphics[width=1\linewidth]{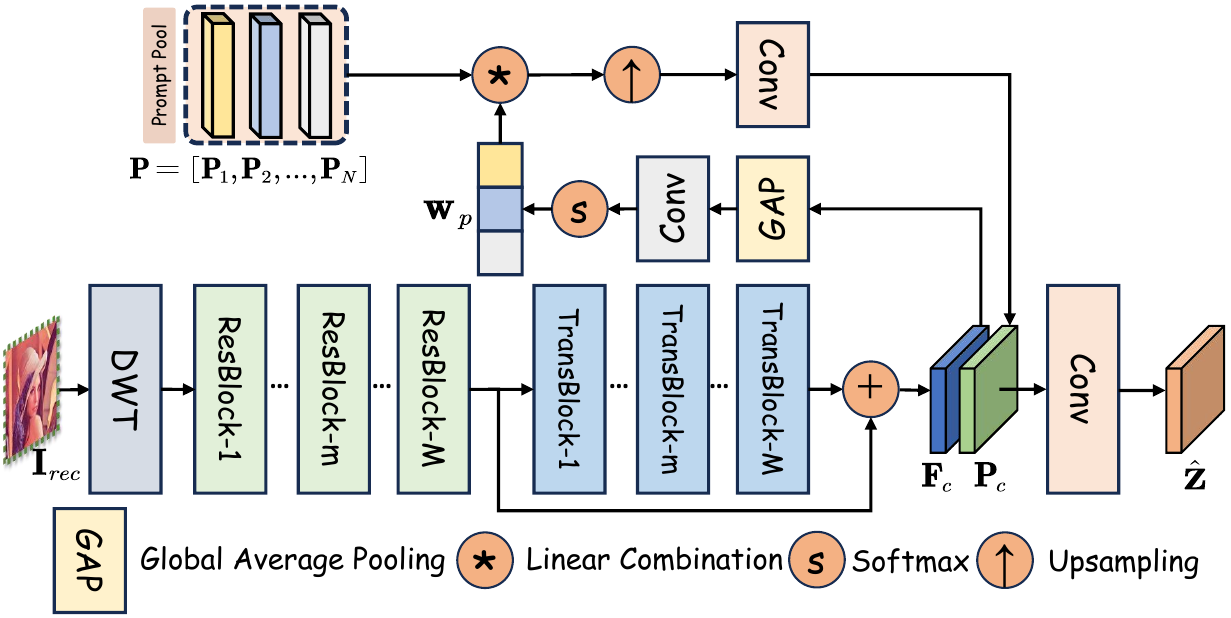}
	\vspace{-10pt}
	\caption{Illustration of the proposed prompt-based posterior estimation. It will dynamically fuse degraded representations and extracted features to obtain posterior mean $\hat{\mathbf{Z}}$$=$$\mathbb{E}[\mathbf{Z}|\mathbf{I}_{rec}]$.}
    \vspace{-20pt}
    \label{ppem}
\end{figure}

To bolster the fidelity and robustness of the image hiding and revealing module, we introduce a degradation prompt-based posteriori estimation module (PPEM). Since the encoding network tends to compress $\left[\mathbf{I}_{ori}; \mathbf{W}_{loc} \right]$$\in$$\mathbb{R}^{H \times W \times 6}$ into the container image $\mathbf{I}_{con}$$\in$$\mathbb{R}^{H \times W \times 3}$, previous methods~\cite{lu2021large, xu2022robust} typically utilized a random Gaussian initialization or an all-zero map at the decoding end to compensate for the lost high-frequency channels. Yet, our observations suggest that discarded information lurks within the edges and textures of the container image. Thus, deploying a dedicated network proves to be a more potent strategy in predicting the posterior mean of the vanished localization watermark information $\hat{\mathbf{Z}}$$=$$\mathbb{E}[\mathbf{Z}|\mathbf{I}_{rec}]$. Specifically, as shown in Fig.~\ref{ppem}, we stack $M$ residual blocks $\operatorname{Res}(\cdot)$~\cite{he2016deep} and $M$ channel-wise transformer blocks $\operatorname{Trans}(\cdot)$~\cite{zamir2022restormer} to extract the local and non-local features $\mathbf{F}_c$. 
{\setlength\abovedisplayskip{0.1cm}
\setlength\belowdisplayskip{0.1cm}
\begin{equation}
\mathbf{F}_c=\operatorname{Trans}(\operatorname{Res}(\operatorname{DWT}(\mathbf{I}_{rec})))+\operatorname{Res}(\operatorname{DWT}(\mathbf{I}_{rec})).
\end{equation}}

Considering that the container image is prone to various degradations during network transmission, we pre-define $N$ learnable embedding tensors as degradation prompts $\mathbf{P} = [\mathbf{P}_1, \mathbf{P}_2,\ldots, \mathbf{P}_N]$, where $N$ denotes the number of degradation types and is set to 3. These learned prompts $\mathbf{P}$ can adaptively learn a diverse range of degradation representations and are integrated with the intrinsic features extracted from $\mathbf{I}_{rec}$, enabling the proposed IBSN to handle multiple types of degradations using a single set of parameters. To better foster the interaction between the input features $\mathbf{F}_c$ and the degradation prompt $\mathbf{P}$, the features $\mathbf{F}_c$ are passed to a global average pooling (GAP) layer, a $1$$ \times$$1$ convolution, and a softmax operator to produce a set of dynamic weight coefficients. Each degradation prompt $\mathbf{P}_i$ is combined using these dynamic coefficients $\mathbf{w}_{p\circledast i}$ and subsequently integrated via an upsampling operator $\uparrow$ and $3$$\times$$3$ convolution to obtain the enhanced representation $\mathbf{P}_c$.
{\setlength\abovedisplayskip{0.15cm}
\setlength\belowdisplayskip{0.15cm}
\begin{equation}
\begin{gathered}
\mathbf{P}_c=\operatorname{Conv}_{3 \times 3}((\sum\nolimits_{i=1}^N \mathbf{w}_{p\circledast i} \mathbf{P}_i)_{\uparrow}), \\
where~~~ \mathbf{w}_p=\operatorname{Softmax}\left(\operatorname{Conv}_{1 \times 1}\left(\text{GAP}\left(\mathbf{F}_c\right)\right)\right).
\end{gathered}
\end{equation}}

Finally, we utilize a $3 \times 3$ convolution to fuse the learned degradation representation with extracted features $\mathbf{F}_c$ to enrich the degradation-specific context, obtaining $\hat{\mathbf{Z}}$.
{\setlength\abovedisplayskip{0.15cm}
\setlength\belowdisplayskip{0.15cm}
\begin{equation}
\hat{\mathbf{Z}}=\operatorname{Conv}_{3 \times 3}\left(\left[\mathbf{P}_c ; \mathbf{F}_c\right]\right) \in \mathbb{R}^{\frac{H}{2} \times \frac{W}{2} \times 12}.
\end{equation}}

\subsection{Bit Encryption and Recovery Modules} 
As shown in Fig.~\ref{framework}, to encode the copyright watermark $\mathbf{w}_{cop}$ into $\mathbf{I}_{med}$, we firstly expand $\mathbf{w}_{cop}$$\in$$\{0, 1\}^L$ via stacked MLPs and reshape it into several $L$$\times$$L$ message feature maps. Meanwhile, $\mathbf{I}_{med}$ is fed to a U-style feature enhancement network to extract features of each downsampling and upsampling layer. Finally, the message features will be upscaled and integrated with multi-level image features via the fusion mechanism~\cite{wu2023sepmark, hu2018squeeze}, achieving the modulation of bit-image information. In the decoding end, $\mathbf{I}_{rec}$ is fed into a U-shaped sub-network and downsampled to a size of $L$$\times$$L$. The recovered copyright watermark $\hat{\mathbf{w}}_{cop}$ is then extracted via an MLP. More details are presented in \textbf{\textcolor{blue}{S.M.}}.
\begin{table}[t!]
\caption{Localization precision (F1-Score) and bit accuracy (BA) comparison with other competitive methods on~\cite{dong2013casia, wen2016coverage, guan2019mfc, hsu2006detecting}.}
\vspace{-0.6cm}
\begin{center}
\resizebox{1.\linewidth}{!}{
\begin{tabular}{c|cc|cc|cc|cc}
\toprule[1.5pt]
\multirow{2}{*}{{Method}}  & \multicolumn{2}{c|}{CAISAv1~\cite{dong2013casia}} & \multicolumn{2}{c|}{Coverage~\cite{wen2016coverage}} & \multicolumn{2}{c|}{NIST16~\cite{guan2019mfc}} & \multicolumn{2}{c}{Columbia~\cite{hsu2006detecting}}   \\ 
 & F1 & BA(\%) & F1 & BA(\%) & F1 & BA(\%) & F1 & BA(\%) \\ \hline
ManTraNet~\cite{wu2019mantra} & 0.320 &- & 0.486 &- & 0.225 &- & 0.650 &-\\
SPAN~\cite{hu2020span} & 0.169 &- & 0.428 &- & 0.363 &- & 0.873 &- \\
CAT-Net v2~\cite{kwon2022learning}   & \underline{0.852} &- & 0.582 &- & 0.417 &- &\underline{0.923} &- \\
OSN~\cite{wu2022robust} &0.676 &- &0.472 &- &0.449 &- &0.836 &- \\ 
MVSS-Net~\cite{dong2022mvss}  & 0.650 &- & 0.659 &- & 0.372 &- & 0.781 &- \\
PSCC-Net~\cite{liu2022pscc} & 0.670 &- & 0.615 &- & 0.210 &- & 0.760 &- \\
TruFor~\cite{guillaro2023trufor} &0.822 &- &\underline{0.735} &- &\underline{0.470} &- &0.914 &- \\
EditGuard (Ours) &\textbf{0.954} & \textbf{99.91} &\textbf{0.955} &\textbf{100} &\textbf{0.911} &\textbf{99.88} &\textbf{0.988} &\textbf{99.93} \\ \bottomrule[1.5pt] 
\end{tabular}}
\label{classical}
\vspace{-15pt}
\end{center}
\end{table}
\begin{figure}[t!]
	\centering
\includegraphics[width=1\linewidth]{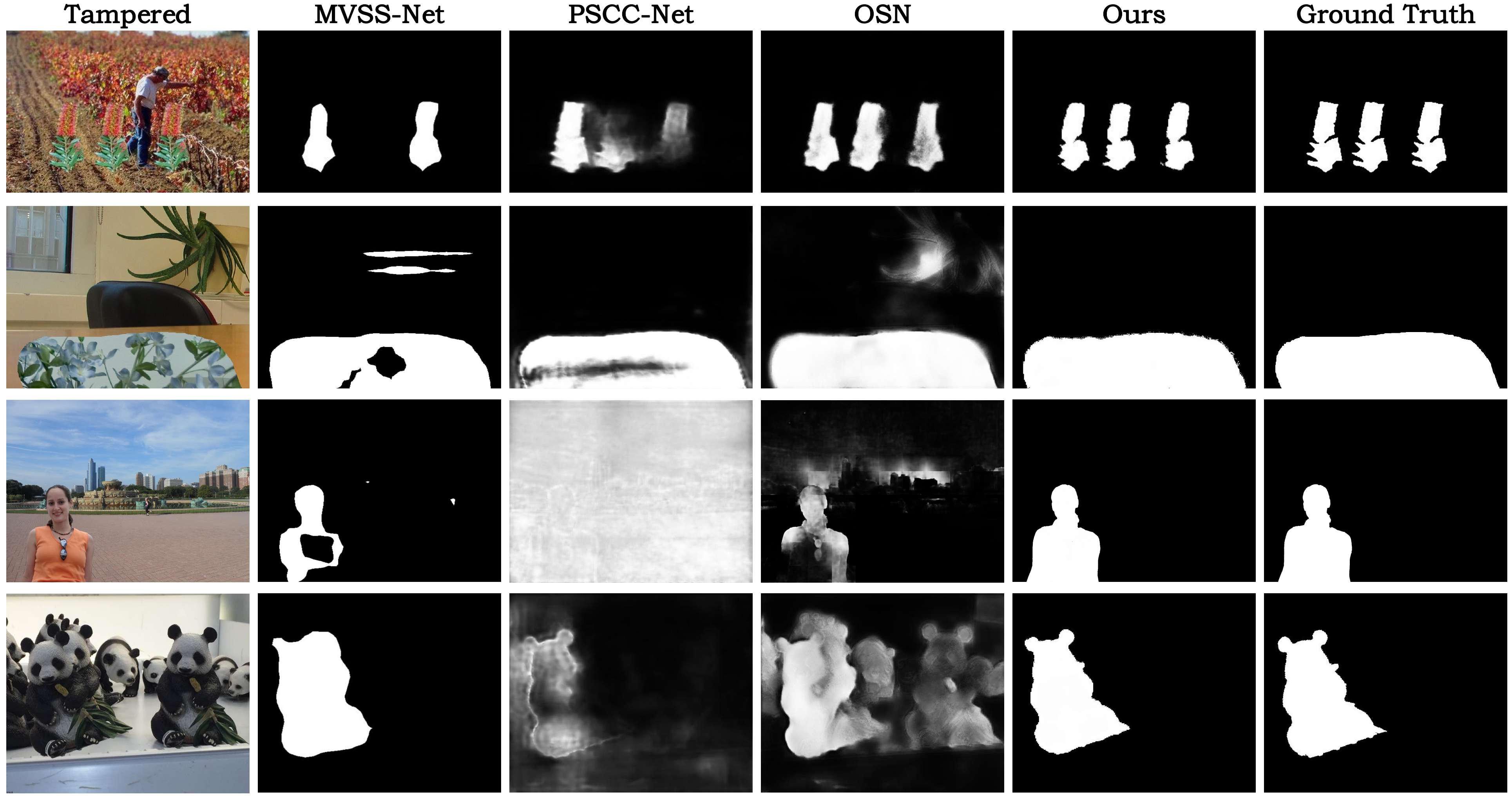}
	\vspace{-18pt}
	\caption{Localization comparisons of our EditGuard and competitive methods~\cite{dong2022mvss, liu2022pscc, wu2022robust} on four classical benchmarks.}
 \vspace{-16pt}
    \label{casia}
\end{figure}
\subsection{Construct EditGuard via the IBSN}
To stabilize the optimization of the proposed IBSN, we propose a bi-level optimization strategy. Given an arbitrary original image $\mathbf{I}_{med}$ and watermark $\mathbf{w}_{cop}$, we first train the bit encryption and recovery module via the $\ell_2$ loss.
{\setlength\abovedisplayskip{0.15cm}
\setlength\belowdisplayskip{0.15cm}
\begin{equation}
    \ell_{cop}=\left\|\mathbf{I}_{con}-\mathbf{I}_m\right\|_2^2+\lambda \left\|\hat{\mathbf{w}}_{cop}-\mathbf{w}_{cop}\right\|_2^2,
\end{equation}}where $\lambda$ is set to $10$. Furthermore, we freeze the weights of BEM and BRM and jointly train the IHM and IRM. Given a random original image $\mathbf{I}_{ori}$, localization watermark $\mathbf{W}_{loc}$ and copyright watermark $\mathbf{w}_{cop}$, the loss function is:
{\setlength\abovedisplayskip{0.15cm}
\setlength\belowdisplayskip{0.15cm}
\begin{equation}
    \footnotesize
    \ell_{loc} = \|\hat{\mathbf{I}}_{ori}-\mathbf{I}_{ori}\|_1 + \alpha\left\|\mathbf{I}_{con}-\mathbf{I}_{ori}\right\|_2^2+\beta\|\hat{\mathbf{W}}_{loc}-\mathbf{W}_{loc}\|_1,
\end{equation}}where $\alpha$ and $\beta$ are respectively set to 100 and 1. During training, we only introduced degradations to $\mathbf{I}_{con}$ without being exposed to any tampering. After acquiring a pre-trained IBSN, we can construct the proposed EditGuard via the components of IBSN. As plotted in Fig.~\ref{framework}, the dual-watermark encoder of EditGuard is composed of IHM and BEM, which correspond to the localization and copyright watermarking in Fig.~\ref{editguard} respectively. The copyright extractor strictly corresponds to BRM. The tamper locater includes IRM and a mask extractor (ME). Note that we need to pre-define a localization watermark $\mathbf{W}_{loc}$, which is shared between the encoding and decoding ends. The choice of $\mathbf{W}_{loc}$ is very general to our method. It can be any natural image or even a solid color image. Finally, by comparing the pre-defined watermark $\mathbf{W}_{loc}$ with the decoded one $\hat{\mathbf{W}}_{loc}$, we can obtain a binary mask $\hat{\mathbf{M}}$$\in$$\mathbb{R}^{H \times W}$:
{\setlength\abovedisplayskip{0.15cm}
\setlength\belowdisplayskip{0.15cm}
\begin{equation}
\hat{\mathbf{M}}[i, j]=\theta_\tau(\max (|\hat{\mathbf{W}}_{loc}[i, j,:]-\mathbf{W}_{l o c}[i, j,:]|)).
\end{equation}}where $i \in [0, H)$ and $j \in [0, W)$. $\mathcal{\theta}_\tau(z) = 1$ ($z \geq \tau$). $\tau$ is set to 0.2. $|\cdot|$ is an absolute value operation.


\section{Experiments}
\begin{table}[t!]
\caption{Visual quality of the container image $\mathbf{I}_{con}$ and bit accuracy comparison with other pure watermarking methods. }
\vspace{-0.6cm}
\begin{center}
\resizebox{1.\linewidth}{!}{
\begin{tabular}{c|c|c|c|c|c|c}
\toprule[1.5pt]
Method  &Image Size &M. L. &PSNR (dB) &SSIM &NIQE($\downarrow$) & BA(\%)   \\ \hline
MBRS~\cite{jia2021mbrs} & 128$\times$128 & 30 &26.57 &0.886 &7.219 & \textbf{100} \\
CIN~\cite{ma2022towards} & 128$\times$128 & 30 &\textbf{41.35} &\textbf{0.981} &7.171 & \underline{99.99}\\ 
PIMoG~\cite{fang2022pimog} & 128$\times$128 & 30 &36.22 &0.941 &7.113 & \underline{99.99} \\ 
SepMark~\cite{wu2023sepmark} & 128$\times$128 & 30 &35.42  &0.931 &\underline{7.095} &99.86 \\
EditGuard &128$\times$128 & 30 &\underline{36.93} &\underline{0.944} &\textbf{5.567} &99.89 \\
EditGuard &512$\times$512 & 64 &37.77 &0.949 &4.257 & 99.95 \\ \bottomrule[1.5pt] 
\end{tabular}}
\label{visual}
\vspace{-28pt}
\end{center}
\end{table}

\subsection{Implementation Details}
We trained our EditGuard via the training set of COCO~\cite{lin2014microsoft} \textbf{without any tampered data}. Thus, for tamper localization, our method is actually zero-shot. The Adam~\cite{kingma2014adam} is used for training 250$K$ iterations with $\beta_1$$=$$0.9$ and $\beta_2$$=$$0.5$. The learning rate is initialized to $1$$\times$$10^{-4}$ and decreases by half for every 30$K$ iterations, with the batch size set to 4. We embed a 64-bit copyright watermark and a simple localization watermark such as a pure blue image ([R, G, B] = [$0$, $0$, $255$]) to original images. Following~\cite{dong2022mvss, guillaro2023trufor, liu2022pscc}, F1-score, AUC, IoU, and bit accuracy are used to evaluate localization and copyright protection performance. Since no prior methods can simultaneously achieve this dual forensics, we conducted separate comparisons with tamper localization and image watermarking methods.

\begin{table*}[ht]
\caption{Comparison with other competitive tamper localization methods under different AIGC-based editing methods. Note that $\dagger$ denotes the network finetuned in our constructed AGE-Set-C. }
\vspace{-10pt}
\resizebox{1.\linewidth}{!}{
\begin{tabular}{c|cccc|cccc|cccc|cccc|cccc|cccc}
\toprule[1.5pt]
\multicolumn{1}{c|}{\multirow{2}{*}{Method}} & \multicolumn{4}{c|}{Stable Diffusion Inpaint~\cite{rombach2022high}} & \multicolumn{4}{c|}{Controlnet~\cite{zhang2023adding}} & \multicolumn{4}{c|}{SDXL~\cite{podell2023sdxl}} & \multicolumn{4}{c|}{RePaint~\cite{lugmayr2022repaint}} & \multicolumn{4}{c|}{Lama~\cite{suvorov2022resolution}} & \multicolumn{4}{c}{FaceSwap~\cite{faceswap}}          \\ 
 & F1 & AUC & IoU & BA(\%) & F1 & AUC & IoU & BA(\%) & F1 & AUC & IoU & BA(\%) & F1 & AUC & IoU & BA(\%) & F1 & AUC & IoU & BA(\%) & F1 & AUC & IoU & BA(\%)\\ \hline
MVSS-Net~\cite{dong2022mvss}  &0.178 & 0.488 & 0.103 &- &0.178 &0.492 &0.103 &- &0.037 &0.503 &0.028 &- &0.104 &0.546 &0.082 &- &0.024 &0.505 &0.022 &- &0.285 &0.612 &0.192 &- \\
OSN~\cite{wu2022robust} &0.174 &0.486 &0.101 &- &0.191 &0.644 &0.110 &- &0.200 &0.755 &0.118 &- &0.183 &0.644 &0.105 &- &0.170 &0.430 &0.099 &- &0.308 &0.791 &0.171 &- \\
PSCC-Net~\cite{liu2022pscc} & 0.166 & 0.501 &0.112 &- &0.177 &0.565 &0.116 &- &0.189 &0.704 &0.115 &- &0.140 &0.469 &0.109 &- &0.132 &0.329 &0.104 &- &0.157 &0.346 &0.180 &- \\
IML-VIT~\cite{ma2023iml}  &0.213 &0.596 &0.135 &- &0.200 &0.576 &0.128 &- &0.221 &0.603 &0.145 &- &0.103 &0.497 &0.059 &- &0.105 &0.465 &0.064 &- &0.105 &0.465 &0.064 &- \\
HiFi-Net~\cite{guo2023hierarchical} &0.547 &0.734 &0.128 &- &0.542 &0.735 &0.123 &- &\underline{0.633} &0.828 &0.261 &- &\underline{0.681} &\underline{0.896} &\underline{0.339} &- &\underline{0.483} &0.721 &0.029 &- &\underline{0.781} &\underline{0.890} &\underline{0.478} &- \\
$\text{MVSS-Net}^{\dagger}$~\cite{dong2022mvss} &\underline{0.694} &\underline{0.939} &\underline{0.575} &- &\underline{0.678} &\underline{0.925} &\underline{0.558} &- &0.482 &\underline{0.884}  &\underline{0.359} &- &0.185 &0.529 &0.111 &- &0.393 &\underline{0.829} &\underline{0.275} &- &0.459 &0.739 &0.333 &- \\
EditGuard (Ours) & \textbf{0.966} &\textbf{0.971} &\textbf{0.936} &\textbf{99.95} &\textbf{0.968} &\textbf{0.987} &\textbf{0.940} &\textbf{99.96} &\textbf{0.965} &\textbf{0.989} &\textbf{0.936} &\textbf{99.96} &\textbf{0.967} &\textbf{0.977} &\textbf{0.938} &\textbf{99.95} &\textbf{0.965} &\textbf{0.969} &\textbf{0.934} &\textbf{99.95} &\textbf{0.896} &\textbf{0.943} &\textbf{0.876} &\textbf{99.86} \\ \bottomrule[1.5pt]
\end{tabular}}
\vspace{-15pt}
\label{age-c}
\end{table*}

\subsection{Comparison with Localization Methods}

For a fair comparison with tamper localization methods, we conducted extensive evaluations on four classical benchmarks~\cite{dong2013casia, wen2016coverage, guan2019mfc, hsu2006detecting}, as reported on Tab.~\ref{classical}. Since EditGuard is a proactive approach, we initially embed watermarks into authentic images and then paste the tampered areas into the container images. Remarkably, even for tamper types that existing methods specialize in, the localization accuracy of EditGuard consistently outperforms the SOTA method~\cite{guillaro2023trufor} across four datasets by margins of \textbf{0.102, 0.116, 0.441, and 0.065 in F1-score without any labeled data or tampered samples required}, which verifies the superiority of our proactive localization mechanism. As shown in Fig.~\ref{casia}, our EditGuard can precisely pinpoint pixel-level tampered areas but other methods can only produce a rough outline or are only effective in some cases. Meanwhile, our bit accuracy remains over \textbf{99.8\%} while all other methods can not realize effective copyright protection.
\begin{figure}[t!]
	\centering
	\includegraphics[width=1\linewidth]{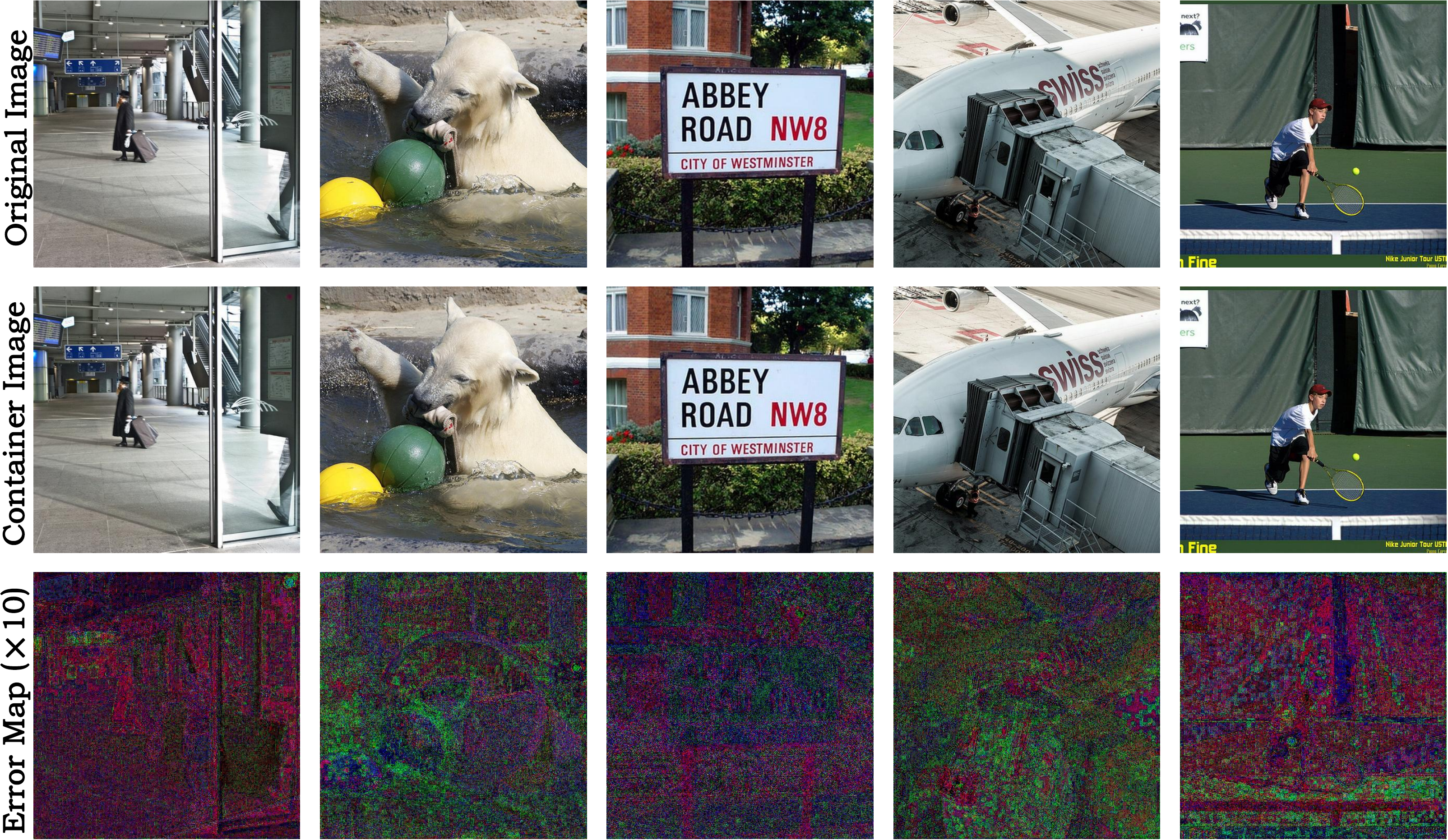}
	\vspace{-18pt}
	\caption{Visual results of the container image $\mathbf{I}_{con}$ and the error map of the proposed EditGuard. Here, localization and copyright watermarks are randomly selected from the dataset.}
 \vspace{-19pt}
    \label{visual1}
\end{figure}
\subsection{Comparison with Watermarking Methods}
To evaluate the visual quality of $\mathbf{I}_{con}$, we compared EditGuard with other watermarking methods on 1$K$ testing images from COCO~\cite{lin2014microsoft} under the tampering of stable diffusion inpaint~\cite{rombach2022high}. For a fair comparison, we also retrained our EditGuard on 128$\times$128 original images and 30 bits. Tab.~\ref{visual} reports that the fidelity of our container image far surpasses that of SepMark~\cite{wu2023sepmark}, PIMoG~\cite{fang2022pimog}, and MBRS~\cite{jia2021mbrs} but is slightly inferior to CIN~\cite{ma2022towards}. Meanwhile, our method exhibits the best performance in perceptual quality measures like NIQE. As shown in Fig.~\ref{visual1}, dual-watermarked images do not have noticeable artifacts and noise, making them imperceptible to the human eyes. When suffer malicious tampering, our method outperforms SepMark and is very close to PIMoG and CIN in bit accuracy. Note that other competitive methods only hide 30 bits, with a capacity of 30/(128$\times$128). In contrast, our EditGuard hides both an RGB localization watermark and a 1D copyright watermark, with a capacity far greater than 30/(128$\times$128). Here, we do not claim to achieve the best visual quality and bit accuracy, but just to demonstrate that our method is comparable to the current image watermarking methods.

\subsection{Extension to AIGC-based Editing Methods}
\textbf{Dataset Preparation:} 
We constructed a dataset tailored for AIGC Editing methods, dubbed AGE-Set, comprising two sub-datasets. The first AGE-Set-C is a batch-processed coarse tamper dataset. Its original images are sourced from COCO 2017~\cite{lin2014microsoft} and CelebA~\cite{liu2015deep}, containing 30$K$ training images and 1.2$K$ testing images. We used some SOTA editing methods such as Stable Diffusion Inpaint~\cite{rombach2022high}, Controlnet~\cite{zhang2023adding}, SDXL~\cite{podell2023sdxl} to manipulate images with the prompt to be ``None'', and employed some unconditional methods like Repaint~\cite{lugmayr2022repaint}, Lama~\cite{suvorov2022resolution}, and Faceswap~\cite{faceswap}. \textbf{Note that we only use the tampered data to train other methods, not our EditGuard.} The second sub-dataset AGE-Set-F includes 100 finely edited images. It is edited manually via some sophisticated software such as SD-Web-UI, Photoshop, and Adobe Firefly. These AIGC-based editing methods can achieve a good fusion of the tampered and unchanged areas, making it hard for the naked eye to catch artifacts. More details are presented in \textcolor{blue}{\textbf{S.M.}}.

\begin{figure}[t!]
	\centering
\includegraphics[width=1\linewidth]{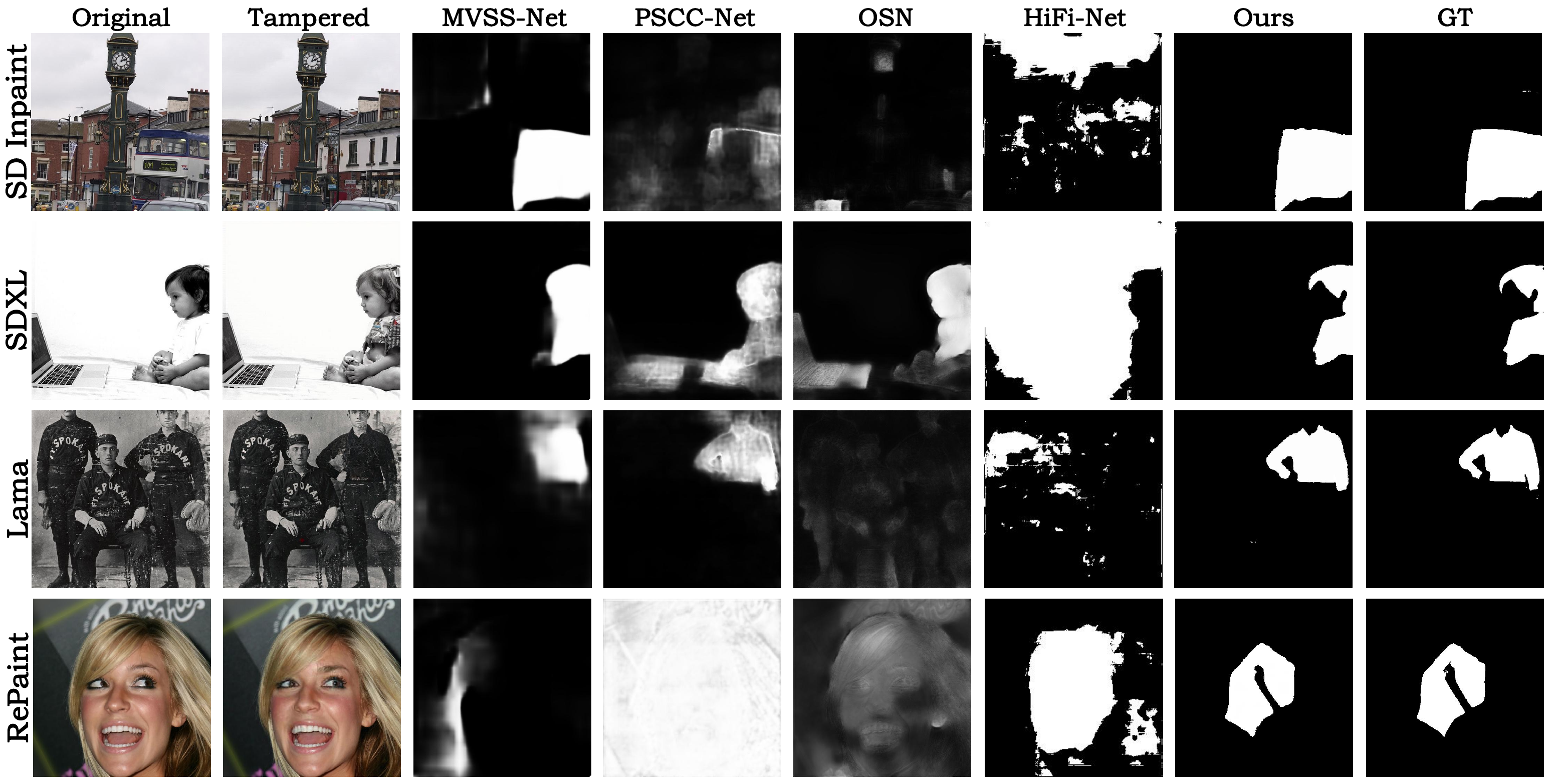}
	\vspace{-20pt}
	\caption{Localization performance comparisons of our EditGuard and other methods~\cite{dong2022mvss, liu2022pscc, wu2022robust, guo2023hierarchical} on our constructed AGE-Set-C.}
 \vspace{-18pt}
    \label{agec}
\end{figure}

\begin{figure*}[t]
	\centering
	\includegraphics[width=1\linewidth]{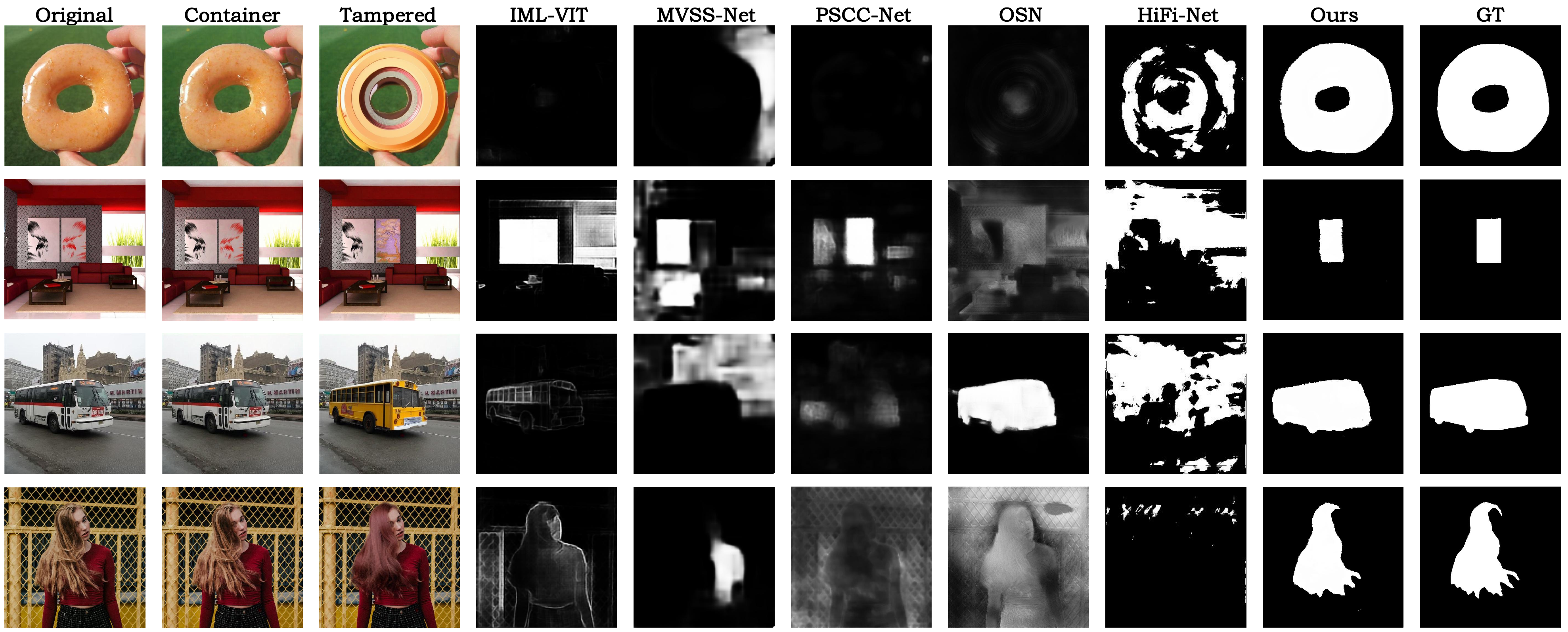}
	\vspace{-22pt}
	\caption{Localization precision comparisons of our EditGuard and other competitive methods on the meticulously tampered AGE-Set-F.}
    \vspace{-16pt}
    \label{AGEF}
\end{figure*}

\textbf{AGE-Set-C:} Tab.~\ref{age-c} presents the comparison of our EditGuard and some SOTA tamper localization methods~\cite{dong2022mvss, wu2022robust, ma2023iml, guo2023hierarchical, liu2022pscc}. We observe that the F1-scores of other passive forensic methods are generally lower than 0.7 when applied to AGE-Set-C. Meanwhile, even when we try our best to finetune MVSS-Net using AGE-Set-C, the accuracy of $\text{MVSS-Net}^{\dagger}$ remains unsatisfactory, and they exhibit catastrophic forgetting across various tamper methods. In contrast, our method can guarantee an F1-score and AUC of over \textbf{95\%}, maintaining around \textbf{90\%} IOU, regardless of tampering types. As shown in Fig.~\ref{agec}, our EditGuard can accurately capture these imperceptible tampering traces produced by AIGC-based editing methods, but other methods are almost ineffective. Moreover, our EditGuard can effectively recover copyright information with a bit accuracy exceeding \textbf{99.8\%}. Noting that none of the comparison methods offer copyright protection capabilities. 

\textbf{AGE-Set-F:} To further highlight the practicality of our EditGuard, we conducted subjective comparisons with other methods on the meticulously tampered AGE-Set-F. The tamper types in this subset did not appear in the training set. As shown in Fig.~\ref{AGEF}, when faced with real-world tampering, even the most powerful tamper localization methods almost entirely fail. This is due to their mechanisms to look for image artifacts and explore instance-wise semantic information. However, our EditGuard, which locates tampered masks via the natural fragility and locality of I2I steganography, can still clearly annotate the tampered area.

\begin{table}[t!]
\caption{Localization and bit recovery performance of our EditGuard and $\text{MVSS-Net}^{\dagger}$~\cite{dong2022mvss} under different levels of degradations.}
\vspace{-8pt}
\resizebox{1.\linewidth}{!}{
\begin{tabular}{c|c|c|cc|ccc|c}
\toprule[1.5pt]
\multirow{2}{*}{Methods} &\multirow{2}{*}{Metrics} & \multirow{2}{*}{Clean} & \multicolumn{2}{c|}{Gaussian Noise} & \multicolumn{3}{c|}{JPEG} & \multirow{2}{*}{Poisson} \\ \cline{4-8} 
                         & &                        & $\sigma$=1             & $\sigma$=5               & $Q$=70    & $Q$=80   & $Q$=90   & \\ \hline
\multirow{2}{*}{$\text{MVSS-Net}^{\dagger}$~\cite{dong2022mvss}}   &F1 & 0.694  & 0.644  & 0.619  & 0.458 & 0.507  & 0.558  & 0.652   \\
&BA(\%) & -  & -  & -  & - & -  & -  & - \\ \hline
\multirow{2}{*}{EditGuard (Ours)}  &F1  & 0.966  & 0.937  & 0.932  & 0.920   & 0.920  & 0.925  & 0.943   \\
     &BA(\%) & 99.95  & 99.94  & 99.37 & 97.69 & 98.16 & 98.23  & 99.91   \\ \bottomrule[1.5pt]
\end{tabular}}
\label{robust}
\vspace{-18pt}
\end{table}

\subsection{Robustness Analysis}
As shown in Tab.~\ref{robust}, we conducted robustness analysis on the tampering of ``Stable Diffusion Inpaint''~\cite{rombach2022high} under Gaussian noise with $\sigma$=1 and 5, JPEG compression with $Q$=70, 80 and 90, and Poisson noise with $\alpha$=4~\cite{zhang2022practical}. We observed that our method still maintains a high localization accuracy (F1-score$\textgreater$0.9) and bit accuracy with a very slight performance decrease under various levels of degradations, while $\text{MVSS-Net}^{\dagger}$~\cite{dong2022mvss} exhibits a noticeable performance degradation compared to its results in clean conditions. It is attributed to our prompt-based estimation that can effectively learn the degradation representation.

\subsection{Ablation Study}
To verify the effectiveness of each component of the EditGuard, we conducted ablation studies on bi-level optimization (BO), lightweight feature interaction module (LFIM), transformer block (TB), and prompt-based fusion (PF) under the tampering of ``Stable Diffusion Inpaint''. As listed in Tab.~\ref{aba}, without BO, the joint training of all components cannot converge effectively, resulting in bit accuracy that is close to random guessing. Without LFIM and TB, the IoU of EditGuard will suffer 0.032 and 0.009 declines since these two modules can better perform feature fusion. Without PF, the robustness of the EditGuard will significantly decline. We observed that the F1/AUC/IoU of our method far surpasses that of case (d) by 0.035/0.031/0.046 under ``Random Degradations'', which indicates that the PF effectively enables a single network to support watermark recovery under various degradations. ``Random Degradations'' denotes that we randomly set the $\mathcal{D}(\cdot)$ to various levels of Gaussian noise, Poisson noise, and JPEG compression.

\begin{table}[t!]
\caption{Abalation studies on the core components of EditGuard.}
\vspace{-8pt}
\resizebox{1.\linewidth}{!}{
\begin{tabular}{c|c|cccc|cccc}
\toprule[1.5pt]
Case &Degradation Type $\mathcal{D}(\cdot)$ & PF & TB & LFIM &BO & F1 & AUC & IoU & BA(\%) \\ \hline
(a)  &Clean & $\checkmark$   & $\checkmark$  & $\checkmark$ &  & -   & -    & -    & 49.17        \\
(b)  &Clean & $\checkmark$  & $\checkmark$  &  &$\checkmark$  &0.950 &0.960  &0.904  &99.73         \\
(c)  &Clean & $\checkmark$  &    & $\checkmark$ &$\checkmark$  &0.957 &0.966  &0.927  &99.51 \\
(d)  &Random Degradations &    & $\checkmark$  & $\checkmark$ & $\checkmark$  &0.903 &0.933  &0.841  &99.12         \\ \hline
\multirow{2}{*}{Ours} &Clean & $\checkmark$  & $\checkmark$  & $\checkmark$ &$\checkmark$ &0.966 &0.971 &0.936 &99.95         \\ 
     &Random Degradations & $\checkmark$  & $\checkmark$  & $\checkmark$ &$\checkmark$ &0.938 &0.964 &0.887 &99.36 \\ \bottomrule[1.5pt]
\end{tabular}}
\vspace{-17pt}
\label{aba}
\end{table}


\section{Conclusion}
\label{sec:conclusion}

We present the first attempt to design a versatile watermarking mechanism \textbf{EditGuard}. It enhances the credibility of images by embedding imperceptible localization and copyright watermarks, and decoding accurate copyright information and tampered areas, making it a reliable tool for artistic creation and legal forensic analysis. In the future, we will focus on improving the robustness of EditGuard and strive not only to offer pixel-wise localization results but also to provide semantic-wise outcomes. Additionally, we plan to further expand EditGuard to a broader range of modalities and applications, including video, audio, and 3D scenes. Our efforts at information authenticity serve not only the AIGC industry, but the trust in our digital world, ensuring that every pixel tells the truth and the rights of each individual are safeguarded.
\clearpage
\maketitlesupplementary

\noindent In the supplementary materials, we demonstrate additional experimental results, implementation details, discussion and analysis, and limitations of our methods as follows.

\tableofcontents
\newpage






\section{More Implementation Details}
\label{sec: editguard}

\subsection{Dataset Preparation}
To construct the AGE-Set-C, over 30$K$ 512$\times$512 original images were selected from the COCO~\cite{lin2014microsoft} and CelebA~\cite{liu2015celeba} datasets and edited using Stable Diffusion Inpaint~\cite{rombach2022high}, ControlNet~\cite{zhang2023adding}, SDXL~\cite{podell2023sdxl}, Repaint~\cite{lugmayr2022repaint}, Lama~\cite{suvorov2022resolution}, and FaceSwap~\cite{faceswap}, where Segment Anything~\cite{kirillov2023segment} was used to extract the tampered regions. Brief introductions of these editing methods are elaborated as follows.

\vspace{3pt}
\noindent \ding{226} Stable Diffusion Inpaint: Stable Diffusion inpainting concatenates the mask with random Gaussian noise as input and performs the diffusion process in the latent domain. It can adaptively generate high-quality image content based on the input mask and prompt, and blend well with unchanged areas.

\vspace{3pt}
\noindent \ding{226} ControlNet: ControlNet inpainting takes the mask and the prompt provided by uses as a condition and injects it into the base Stable Diffusion via an adapter, filling in the missing areas. Note that although they share similar network architectures, ControlNet and Stable Diffusion inpainting are two distinctly different editing methods.

\vspace{3pt}
\noindent \ding{226} SDXL: SDXL inpainting is an advanced version of the Stable Diffusion inpainting, which is trained on multiple scales and incorporates a refinement model to enhance the visual fidelity of generated images.

\vspace{3pt}
\noindent \ding{226} RePaint: RePaint uses a pre-trained diffusion model 
 and only alters the reverse diffusion iterations by sampling the unmasked regions. The prior model is trained on the human face data.

\vspace{3pt}
\noindent \ding{226} Lama: Lama is an unconditional inpainting model and addresses the challenge of inpainting large areas with complex structures by employing Fast Fourier Convolutions (FFCs) and an aggressive mask generation strategy. 

\vspace{3pt}
\noindent \ding{226} FaceSwap: We utilize the open-source project~\cite{faceswap} to swap a random face in CelebA~\cite{liu2015celeba} to another face.
\vspace{3pt}

In a nutshell, we list the settings and division of AGE-Set-C in the Tab.~\ref{settings}. Considering the characteristics of various models, we conducted these four types of tampering~\cite{rombach2022high, zhang2023adding, podell2023sdxl, wu2022robust} on (20$K$ + 1$K$) images from COCO~\cite{lin2014microsoft}, performed two types of tampering ~\cite{faceswap, lugmayr2022repaint} on (10$K$ + 200) human faces respectively. Note that we only use the training set of AGE-Set-C to train passive localization networks such as MVSS-Net~\cite{dong2022mvss}. 

\begin{table*}[h]
\centering
\caption{The settings of the training and testing set of AGE-Set. }
\vspace{-5pt}
\resizebox{!}{1.2cm}{
\begin{tabular}{c|c|c|c}
\toprule[1.5pt]
Division                    & Data Source & Tampering Method                    & Number of Images \\ \hline
\multirow{2}{*}{Training} & COCO        & Stable Diffusion Inpaint, Controlnet, SDXL, Lama & 20$K$×4   \\
                          & Celeba      & RePaint, FaceSwap                   & 10$K$×2   \\ \hline
\multirow{3}{*}{Testing}  & COCO        & Stable Diffusion Inpaint, Controlnet, SDXL, Lama & 1$K$×4    \\
                          & Celeba      & RePaint, FaceSwap                   & 200×2    \\ 
                          & Online      & SD-Web-UI, Adobe firefly, Photoshop & 100    \\
\bottomrule[1.5pt]
\end{tabular}}
\label{settings}
\end{table*}

\subsection{Comparison Methods}
In this section, we will briefly introduce the settings and high-level functions of our comparison methods in Tab.~\ref{setting}. Our comparison methods include two aspects. On the one hand, we compared with many SOTA tamper localization networks to demonstrate that our proactive localization mechanism has extremely outstanding advantages. On the other hand, we compared our EditGuard with the SOTA pure bit-hiding image watermarking method. It shows that we are also close to the best methods in terms of reconstructed bit accuracy and copyright protection.

\begin{table*}[t!]
\centering
\caption{The settings and high-level functions of our comparison methods, where T. L., T.D and C. P. respectively denote tamper localization, tamper detection, and copyright protection.}
\vspace{-10pt}
\resizebox{\linewidth}{!}{
\begin{tabular}{c|c|c|c|c|c}
\toprule[1.5pt]
Method & Reference & Type & Function & Need Tampered Samples & Supported Tampering \\ \hline
MVSS-Net~\cite{dong2022mvss}  & TPAMI 22  & Passive   & T. L. &$\checkmark$ & copy-and-paste, slicing, inpainting   \\ 
OSN~\cite{wu2022robust}       & CVPR 22   & Passive   & T. L. & $\checkmark$   &splicing, removal, copy-and-paste             \\
PSCC-Net~\cite{liu2022pscc}  & TCSVT 22  & Passive   & T. L. &$\checkmark$   &splicing, copy-and-paste, removal, pristine classes             \\
HiFi-Net~\cite{guo2023hierarchical}  & CVPR 23   & Passive   & T. L. &$\checkmark$  &  diffusion, GAN, CNN partial manipulation, copy-and-paste, splicing  \\
IML-VIT~\cite{ma2023iml}   & Arxiv 23  & Passive   & T. L. &$\checkmark$    &splicing, copy-and-paste, inpainting\\
TruFor~\cite{guillaro2023trufor} & CVPR 23 & Passive & T. L. &$\checkmark$  &splicing, copy-and-paste  \\ 
MBRS~\cite{jia2021mbrs}      & ACM MM 21 & Proactive & C. P. &$\times$   &  --             \\
CIN~\cite{ma2022towards}       & ACM MM 22 & Proactive & C. P. &$\times$ & --              \\
PIMoG~\cite{fang2022pimog}    & ACM MM 22 & Proactive & C. P. & $\times$ & -- \\
SepMark~\cite{wu2023sepmark}   & ACM MM 23 & Proactive & T. D. \& C. P. &$\checkmark$ & StarGAN, GANimation, SimSwap \\
EditGuard & -- & Proactive & T. L. \& C. P. &$\times$  & slicing, copy-and-paste, FaceSwap~\cite{faceswap}, AIGC-based editing methods like~\cite{rombach2022high, zhang2023adding, podell2023sdxl, wu2022robust}\\
\bottomrule[1.5pt]
\end{tabular}}
\label{setting}
\end{table*}

\subsection{Details of invertible blocks in IHM and IRM}
\begin{figure*}[t!]
	\centering
	\includegraphics[width=1\linewidth]{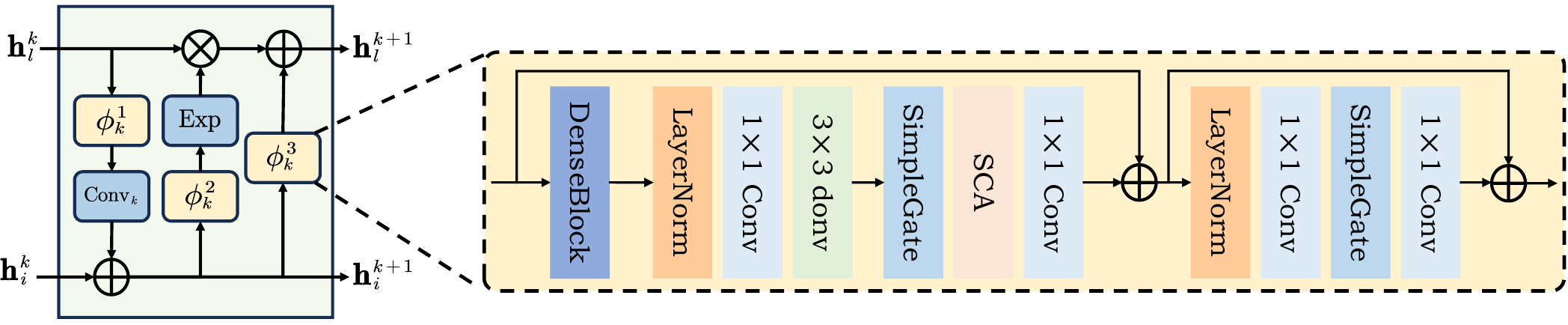}
	\caption{Network structure of our invertible blocks. The enhanced transformation module $\phi_k^i$ is composed of a dense block and a lightweight feature interaction module(LFIM).}
 \vspace{-15pt}
    \label{invblock}
\end{figure*}

To hide and reveal images with high fidelity, we utilize invertible blocks~\cite{lu2021large, mou2023large} in our EditGuard. The original image $\mathbf{I}_{ori}$$\in$$\mathbb{R}^{H \times W\times 3}$ and localization watermark $\mathbf{W}_{loc}$$\in$$\mathbb{R}^{H \times W \times 3}$ undergo discrete wavelet transformations (DWT) to yield frequency-decoupled image features. Subsequently, we employ enhanced additive affine coupling layers to project the features in the original image and its corresponding localization watermark branches. Concretely, in $k$-th invertible block, the bijection of the forward propagation is:
{\setlength\abovedisplayskip{0.15cm}
\setlength\belowdisplayskip{0.15cm}
\begin{equation}
\mathbf{h}_{ori}^{k+1}=\mathbf{h}_{ori}^{k}+\text{Conv}_k\left( \phi _{k}^{1}\left( \mathbf{h}_{loc}^{k} \right) \right),
\label{eq1}
\end{equation}
\begin{equation}
\mathbf{h}_{loc}^{k+1}=\mathbf{h}_{loc}^{k}\otimes \text{Exp}\left( \phi _{k}^{2}\left( \mathbf{h}_{ori}^{k+1} \right) \right) +\phi _{k}^{3}\left( \mathbf{h}_{ori}^{k+1} \right),
\label{eq2}
\end{equation}}where $\mathbf{h}_{ori}^{k}$ and $\mathbf{h}_{loc}^{k}$ 
 respectively denote the feature in the $k$-th layer of the original image and localization watermark branch. $\text{Conv}_k(\cdot)$ and $\text{Exp}(\cdot)$ respectively denote the $3$$\times$$3$ convolution in the $k$-th layer and exponential function. $\phi_k^i(\cdot)\,(i=1, 2, 3)$ denote the enhanced affine coupling layers, which is composed of a five-layer dense convolution block~\cite{mou2023large} and a lightweight feature interaction module (LFIM)~\cite{chen2022simple}. The dense block continuously cascades features extracted from preceding layers, thereby acquiring coarse-grained semantic information. The LFIM can enhance the non-linearity of transformations and capture the long-range dependencies with low computational cost.  As shown in Fig.~\ref{invblock}, the LFIM consists of two-layer normalizations, a simplified channel attention mechanism (SCA), a depth-wise separable $3$$\times$$3$ convolution, four $1$$\times$$1$ convolutions, and a gating function. This lightweight module can better promote feature fusion between channels, capture long-distance dependencies, and filter out redundant information. Correspondingly, the backward propagation process is defined as:
{\setlength\abovedisplayskip{0.15cm}
\setlength\belowdisplayskip{0.15cm}
\begin{equation}
    \mathbf{h}_{loc}^{k}=\left( \mathbf{h}_{loc}^{k+1}-\phi _{k}^{3}\left( \mathbf{h}_{ori}^{k+1} \right) \right) \otimes \text{Exp}\left( -\phi _{k}^{2}\left( \mathbf{h}_{ori}^{k+1} \right) \right),
\label{eq3}
\end{equation}
\vspace{-8pt}
\begin{equation}
\mathbf{h}_{ori}^{k}=\mathbf{h}_{ori}^{k+1}-\text{Conv}_k\left( \phi _{k}^{1}\left( \mathbf{h}_{loc}^{k} \right) \right).
\label{eq4}
\end{equation}}The revealed features are then transformed to the image domain via the inverse discrete wavelet transform (IDWT).

\subsection{Details of BEM and BRM}
To encode the copyright watermark $\mathbf{w}_{cop}$ into $\mathbf{I}_{med}$, we firstly convert the copyright watermark $\mathbf{w}_{cop} \in \{0, 1\}^{L}$ into $\{-0.5, 0.5\}^{L}$ and feed it into stacked MLPs, and reshape it into several message feature maps $\{\mathbf{M}_i\}_{i=1}^N \in \mathbb{R}^{L \times L}$. Meanwhile, $\mathbf{I}_{med}$ is fed to a U-style feature
enhancement network with $N$ layers. In the downsampling, a convolution with stride=2 followed by a ``Conv-ReLU'' layer is used to halve the spatial resolution and double the feature channels, yielding the results $\{\mathbf{D}_i\}_{i=1}^N$. In the upsampling, we employ nearest neighbor interpolation combined with ``Conv-ReLU'' layers to produce features $\{\mathbf{U}_i\}_{i=N}^1$. Subsequently, the message features $\{\mathbf{M}_i\}_{i=1}^N$, derived from several MLPs, are up-scaled via nearest interpolation operation $\uparrow$ and integrated with $\{\mathbf{D}_i\}_{i=1}^N$ and $\{\mathbf{U}_i\}_{i=N}^1$, prompting the modulation of bit-image information.
\begin{equation}
    \hat{\mathbf{U}}_i = \text{Fuse}([\mathbf{D}_i; \mathbf{U}_i; (\mathbf{M}_i)_{\uparrow}]) \in \mathbb{R}^{\frac{H}{2^i} \times \frac{W}{2^i} \times 2^{i}C}
\end{equation}
where \(\text{Fuse}(\cdot)\) denote the residual blocks with channel attention module \cite{wu2023sepmark, hu2018squeeze}. $[;]$ denotes the concat operation. Finally, the BEM combines $\mathbf{I}_{med}$ and $\mathbf{w}_{cop}$ at multiple levels to produce a dual-encoded container image $\mathbf{I}_{con}$. In the decoding process, the received image $\mathbf{I}_{rec}$ is fed into a U-shaped sub-network, subsequently downsampled to a size of $L \times L$. The recovered copyright watermark $\hat{\mathbf{w}}_{cop}$ is then extracted via an MLP and uses $0$ as the threshold to transform it into $\{0, 1\}^{L}$.

\begin{figure*}[t!]
	\centering
	\includegraphics[width=1\linewidth]{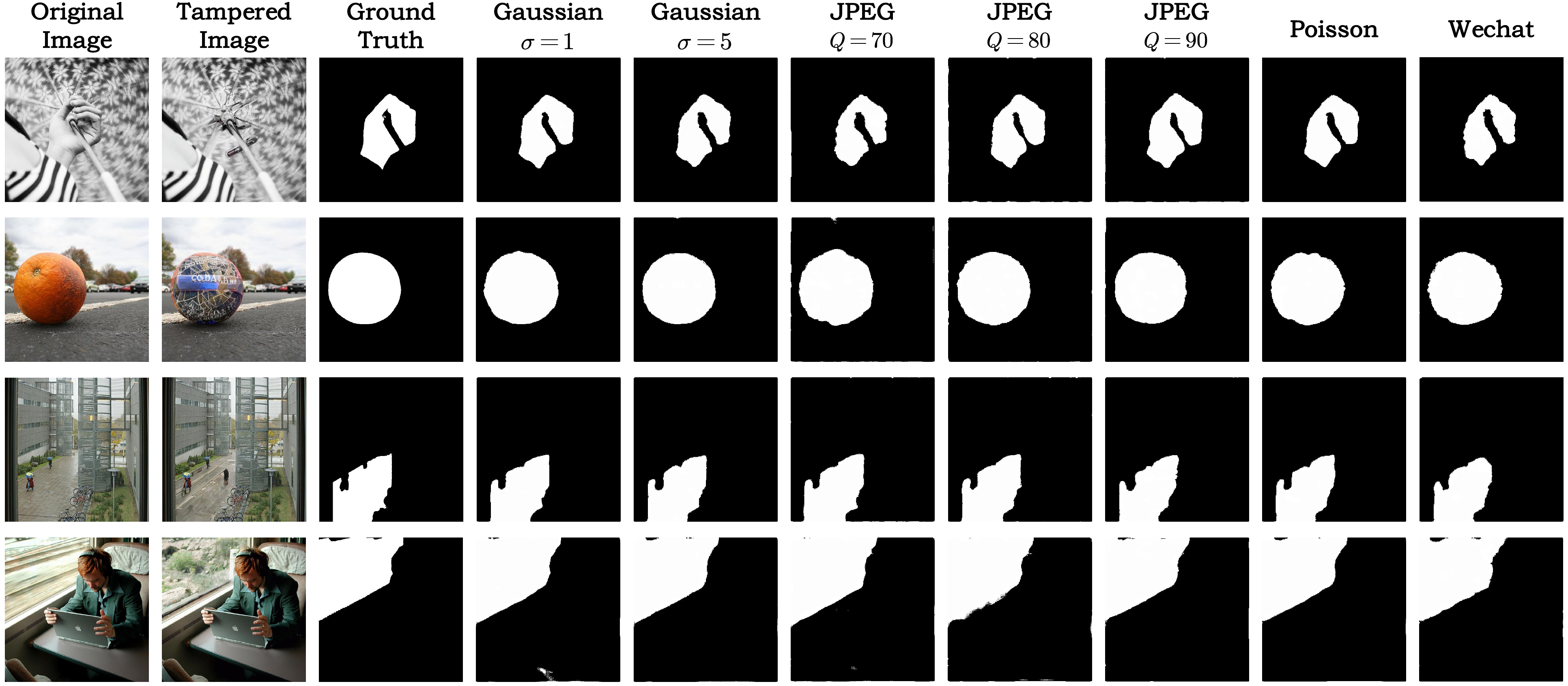}
        \vspace{-20pt}
	\caption{Localization results of our EditGuard on different simulation degradation and ``WeChat'' degradation. Zoom in for a better view.}
 \vspace{-15pt}
    \label{robust1}
\end{figure*}

\subsection{Training Details and Hyperparameters}
The number of invertible blocks in the image hiding and revealing modules is $6$. The number of residual convolution blocks and transformer blocks $M$ is respectively set to $8$. The degradation prompt $\mathbf{P} \in \mathbb{R}^{C \times L \times H \times W}$ is a pre-defined learnable tensor, where $C$, $L$, $H$ and $W$ are respectively set to $72$, $3$, $36$ and $36$. $L$ corresponds to the number of degradation types introduced in our training process. The number of U-shaped feature enhancement layers $N$ is $3$. We introduced a JPEG simulator to train our EditGuard similar to~\cite{xu2022robust}, which consists of four main steps: color space transformation, discrete cosine transformation (DCT), quantization, and entropy encoding. To consider the ability of our model to hide extreme images (solid color images), we feed a pure blue image as a localization watermark every 10$K$ iteration during training. 

\section{Addtional Experiments}
\label{sec: expr}

\subsection{Time and Computational Cost}
To evaluate the computational efficiency and complexity of our method, we input a $3\times512\times512$ tensor on an NVIDIA 3090Ti to test our approach and various tamper localization methods~\cite{dong2022mvss, wu2022robust, ma2023iml, guo2023hierarchical} in terms of inference time, the number of parameters, and FLOPs. We observe from Tab.~\ref{time} that our method boasts the lowest number of parameters and the second lowest inference time and FLOPS, which demonstrates the computational efficiency and minimal demand on computational resources of our EditGuard compared to other methods. This can be attributed to the symmetrical network structure of invertible neural networks, which can save a significant portion of parameters. Note that the inference time of EditGuard here only includes the time taken at the decoding end to obtain the mask and copyright information.
\begin{table}[h]
\centering
\vspace{-5pt}
\caption{Inference time, the number of parameters, and FLOPS comparison of our EditGuard and other methods.}
\vspace{-10pt}
\resizebox{\linewidth}{!}{
\begin{tabular}{c|ccc}
\toprule[1.5pt]
Methods  & Inference Time (s) & \# Params.(M) & FLOPS (G) \\ \hline
MVSS-Net~\cite{dong2022mvss} & 2.929  & 142.78 & 163.57    \\
OSN~\cite{wu2022robust}      & 0.062 & 128.82 & 161.07    \\
IML-VIT~\cite{ma2023iml}     & 0.094 & 88.63  & 445.33    \\
HiFi-Net~\cite{guo2023hierarchical} & 1.512   & 10.13  & 82.27     \\  
EditGuard &0.069  &5.45  &102.73 \\ \bottomrule[1.5pt]
\end{tabular}}
\label{time}
\end{table}

\subsection{More Results of Robustness}
To further showcase the exceptional robustness of our method, we present the visualized localization results of our EditGuard under various degradation, including simulated degradation like Gaussian noise, JPEG compression, Poisson noise, as well as real-world degradation. Specifically, we also display the results of our method under ``WeChat'' degradation. We send and receive tampered container images via the pipeline of WeChat to implement network transmission. As illustrated in Fig.~\ref{robust1}, our EditGuard continues to display accurate localization results in comparison to the ground truth under different degradation. With the increase in degradation levels, it only shows a slight decrease in localization precision, which demonstrates the practicality and robustness of our method.

\subsection{Ablation Study of Localization Watermark}
To investigate the impact of different localization watermarks on the performance of our method, we embedded various localization watermarks into the original image and tampered with them using Stable Diffusion inpaint, exploring the localization accuracy and copyright recovery precision of our EditGuard. As shown in Tab.~\ref{watermark}, our method is applicable to all choices of localization watermark images, demonstrating good generalizability. Meanwhile, our EditGuard performs better on images with simpler textures and slightly worse on randomly selected natural images. Compared with pure red and pure green images, thanks to the fact that we added some pure blue images to the training samples, the localization accuracy of the pure blue watermark is higher, which shows the effectiveness of our training approach. Additionally, we observe that although using a natural image as a localization watermark can effectively reflect the tampered areas, it is challenging to select a suitable threshold $\tau$ for the mask extractor, resulting in relatively low localization accuracy. Therefore, we recommend using a relatively simple localization watermark to enhance localization performance.
\begin{table}[h]
\centering
\vspace{-5pt}
\caption{Localization and bit accuracy performance of different localization watermarks.}
\vspace{-10pt}
\resizebox{\linewidth}{!}{
\begin{tabular}{c|cccc}
\toprule[1.5pt]
Localization Watermark & F1 & AUC & IoU & BA(\%) \\ \hline
Pure red image         &0.946 &0.965 &0.923 &99.92 \\
Pure green image       &0.885 &0.922 &0.838 &99.94  \\
Natural image          &0.851 &0.944 &0.750 &99.97  \\
Pure blue image (Ours) &0.966 &0.971 &0.936 &99.95  \\ \bottomrule[1.5pt]
\end{tabular}}
\label{watermark}
\end{table}

\subsection{Comparison with other Image-into-Image Steganography Methods}
To validate the superiority of our designed image-bit steganography network (IBSN) compared to other image-into-image steganography networks, we conducted experiments under the tampering of Stable Diffusion inpaint with SOTA image-into-image steganography works, HiNet~\cite{jing2021hinet} and RIIS~\cite{xu2022robust}. Firstly, it is important to note that these steganography networks do not possess the capability to hide 2D and 1D watermarks simultaneously and provide copyright protection. To fairly compare localization performance with other methods, we also embedded a pure blue image as a localization watermark into both RIIS and HiNet. We observed that while the localization accuracy of HiNet is acceptable in clean conditions, it completely deteriorates with even a slight degradation. Meanwhile, RIIS sacrifices too much reconstruction fidelity for secret information in order to ensure robustness. Therefore, although it suffers a small performance decrease when faced with degradation, the localization accuracy of RIIS is still much lower than our EditGuard. In contrast, our EditGuard can simultaneously maintain high fidelity of the container image, the ability to hide both 1D and 2D watermarks, high accuracy of secret information recovery, and decent robustness. This makes it well-suited for the task of proactive localization and copyright protection.

\begin{table}[h]
\centering
\vspace{-5pt}
\caption{Localization and bit accuracy performance of our method and other image-into-image steganography methods.}
\vspace{-10pt}
\resizebox{\linewidth}{!}{
\begin{tabular}{c|c|ccccc}
\toprule[1.5pt]
Methods &Depredation Type & F1 & AUC & IoU & BA(\%) \\ \hline
\multirow{2}{*}{HiNet~\cite{jing2021hinet}}  &Clean &0.952 &0.962 &0.911 &- \\
&JPEG(Q=80) &0.025 &0.473 &0.013 &- \\
\multirow{2}{*}{RIIS~\cite{xu2022robust}} &Clean &0.854 &0.901 &0.821 &-   \\
&JPEG(Q=80) &0.833 &0.874 &0.802 &- \\ \hline
\multirow{2}{*}{EditGuard (Ours)} &Clean &0.966 &0.971 &0.936  &99.95    \\ 
&JPEG(Q=80) &0.920 &0.949 &0.869 &98.16 \\ \bottomrule[1.5pt]
\end{tabular}}
\end{table}

\subsection{Security Analysis}
\begin{figure}[t!]
	\centering
	\includegraphics[width=1.\linewidth]{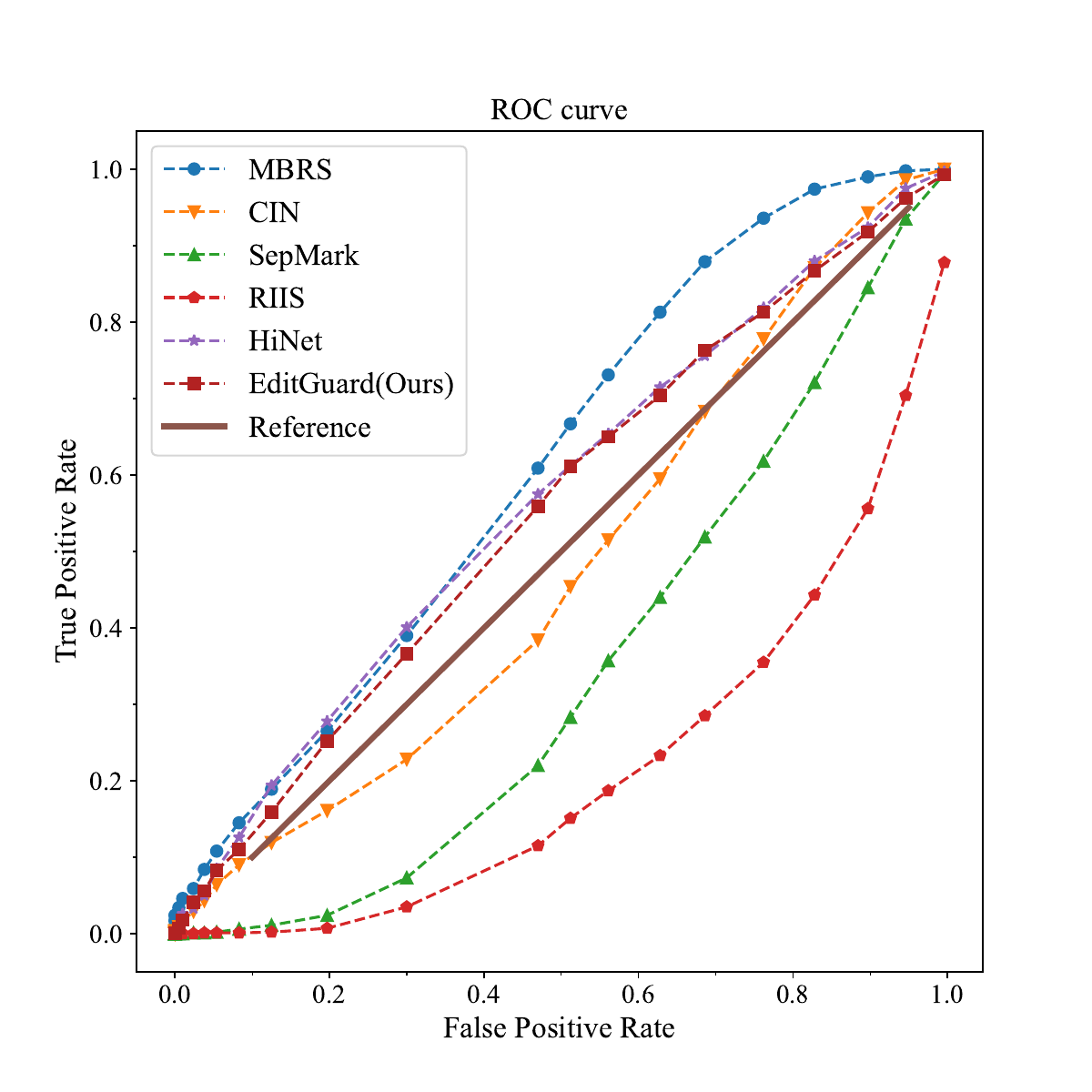}
    \vspace{-15pt}
	\caption{ROC curve of different methods under steganography detector. The closer the curve is to the reference central axis (which means random guess), the corresponding method is better in security.}
 \vspace{-5pt}
    \label{roc}
\end{figure}

To verify the security of our EditGuard, we perform anti-steganography detection via StegExpose~\cite{boehm2014stegexpose} on container images of various methods, including MBRS~\cite{jia2021mbrs}, CIN~\cite{ma2022towards}, SepMark~\cite{wu2023sepmark}, RIIS~\cite{xu2022robust}, HiNet~\cite{jing2021hinet} and our EditGuard. 
Among them, MBRS, CIN, and SepMark embedded a mere 30-bit copyright watermark, while RIIS and HiNet concealed a fragile watermark within an RGB image. In contrast, our EditGuard embedded both a 2D RGB localization watermark and a 1D copyright watermark. The detection set is built by mixing container images and original images with equal proportions. We vary the detection thresholds in a wide range in StegExpose~\cite{boehm2014stegexpose} and draw the receiver operating characteristic (ROC) curve in Fig.~\ref{roc}. Note that the ideal case represents that the detector has a 50\% probability of detecting stego from an equally mixed detection test, the same as a random guess. Evidently, the security of our method exhibits a significant advantage compared to some I2I steganography methods like HiNet and RIIS. It also surpasses most image watermarking methods such as SepMark and MBRS, only slightly lower than CIN. It fully demonstrates that our method is not easily detectable by steganalysis methods, ensuring a high level of security.

\section{Discussion and Analysis}
\label{sec: discuss}

\subsection{Related Works about Image-into-Image (I2I) Steganography}
Image-into-Image (I2I) steganography~\cite{baluja2017hiding} aims to hide a cover image into the host image to produce a container image. Traditional steganography utilized spatial-based methods~\cite{pan2011image, tsai2009reversible, tsai2009reversible, nguyen2006multi, niimi2002high}, adaptive methods~\cite{pevny2010using,li2014new} and transform domain-based schemes~\cite{chanu2012image,kadhim2019comprehensive} to perform hiding and recovery. However, these methods often fail to offer high payload capacity and security. Recently, various deep learning-based steganography approaches have been proposed. Baluja~\cite{baluja2017hiding} firstly proposed to hide a full-size image into another image via a deep network. Generative adversarial schemes~\cite{shi2017ssgan, yang2019embedding, tang2019cnn, zhang2019steganogan, abdal2019image2stylegan, yu2023freedom, yu2023cross} are also introduced to synthesize container images or minimize distortion. Owing to the intrinsic capability of reversible architecture for lossless information retrieval, invertible neural networks~\cite{xiao2020invertible, lu2021large, jing2021hinet, xu2022robust, mou2023large} have been applied to information hiding, greatly enhancing the capacity, visual quality, and fidelity of image steganography system.

\subsection{Relationship to other Tamper Localization Works}
Firstly, we want to reiterate the definitions of proactive and passive tamper localization. \underline{\textbf{Passive tamper localization}} aims to identify and locate tampered areas in an image by analyzing specific features or anomalies, without any preventive measures for the original images before foreseeable manipulations~\cite{wu2023sepmark}. This approach often relies on detecting inconsistencies, unnatural artifacts, or areas that do not match with the surrounding environment in the image. It is entirely based on the analysis of the existing image content. \underline{\textbf{Proactive tamper localization}} refers to the approach of preemptively identifying and pinpointing tampered regions in an image by utilizing pre-embedded signals or markers. Unlike passive methods that rely solely on post-hoc analysis of image anomalies, proactive localization involves the prior integration of specific features, such as image watermarks, which can be subsequently detected and analyzed to reveal any alterations. Although previous efforts have also embedded signals into images for proactive tampering localization, we must emphasize that their applications and approaches are completely orthogonal and distinct from our methods.

\textbf{Comparison with DRAW~\cite{hu2023draw}:} DRAW aims to embed watermarks on RAW images and make passive tamper localization networks more resistant to some lossy image operations such as JPEG compression, blurring, and re-scaling. Therefore, different from our proactive localization mechanism, DRAW fundamentally remains a passive localization method.

\textbf{Comparison with Imuge+~\cite{ying2023learning}:} Imuge+ is employed to embed slight perturbations into the original images for immunization and auto-recover the tampered contents via an invertible network. However, it still uses a passive tamper localization network to extract the masks.

\textbf{Comparison with MaLP~\cite{asnani2023malp}:} 
Although MaLP employs a template matching approach to embed learnable templates into the network, a significant number of tampered samples are still required during the template and network learning process. In contrast, our method does not require any tampered samples during training and is a zero-shot localization method. Meanwhile, the template used by MaLP cannot be changed once it is learned, which results in the image input to the network being only of fixed resolution. But our EditGuard is naturally applicable to images of arbitrary resolution.

In a nutshell, to our knowledge, using a proactive mechanism for tamper localization tasks without the need for tampered training samples remains a scarcely studied issue. Meanwhile, we also believe that the aforementioned excellent works~\cite{hu2023draw, asnani2023malp, ying2023learning} could be effectively combined with our method to achieve better robustness or self-recovery capabilities.

\subsection{Why Choose a Sequential Encoder and Parallel Decoder?}

Apart from sequentially embedding localization and copyright watermarks into original images, a more intuitive idea is to hide a 2D image and a 1D watermark in parallel. Specifically, we first reshape $64$ bits into an $8 \times 8$ image, which is then replicated to match the size of the original image. Subsequently, we retrained a steganography network capable of hiding multiple images within a single image, embedding both the localization watermark and the bit image in parallel into the original image. The localization accuracy and copyright recovery precision are reported in Tab.~\ref{sequential}. We found that the parallel encoding approach performs very well when no degradation is added. However, once even slight degradation is introduced, its bit accuracy dramatically decreases, even approaching 50\% under the JPEG compression ($Q=70$), which is the same with random guessing. This is due to the fact that treating bits entirely as an image and parallel embedding them into the original images leads to local and fragile hiding, making it challenging to achieve adequate robustness.

Meanwhile, the structure of sequential encoding and parallel decoding allows us to flexibly train the image hiding-revealing and bit encryption-recovery branches, enabling these two branches to avoid interference with each other. Considering the different hiding capacities of the two branches, we can even introduce different degradation layers and robustness levels to them, making the overall network more adaptable and flexible. These experimental results and analysis demonstrate the rationality and significance of the framework design of our EditGuard.

\begin{table}[ht]
\centering
\caption{The comparison results of the localization and bit recovery performance between our employed sequential encoding and its counterpart, parallel encoding under the tampering of Stable Diffusion inpaint.}
\vspace{-5pt}
\resizebox{\linewidth}{!}{
\begin{tabular}{c|c|c|cc|ccc|c}
\toprule[1.5pt]
\multirow{2}{*}{Methods} &\multirow{2}{*}{Metrics} & \multirow{2}{*}{Clean} & \multicolumn{2}{c|}{Gaussian Noise} & \multicolumn{3}{c|}{JPEG} & \multirow{2}{*}{Poisson} \\ \cline{4-8} 
                         & &                        & $\sigma$=1             & $\sigma$=5               & $Q$=70    & $Q$=80   & $Q$=90   & \\ \hline
\multirow{2}{*}{Parallel Encoding}  & F1 &0.979	&0.953	&0.911	&0.852 &0.866 &0.879 &0.941 \\
&BA(\%) &100.00 &85.29 &70.28 &58.12 &65.24 &70.52 &79.29 \\ \hline
\multirow{2}{*}{EditGuard (Ours)} & F1 & 0.966  & 0.937  & 0.932  & 0.920   & 0.920  & 0.925  & 0.943   \\
&BA(\%) & 99.95  & 99.94  & 99.37 & 97.69 & 98.16 & 98.23  & 99.91   \\ \bottomrule[1.5pt]
\end{tabular}}
\label{sequential}
\end{table}

\begin{figure*}[ht]
	\centering
	\includegraphics[width=1\linewidth]{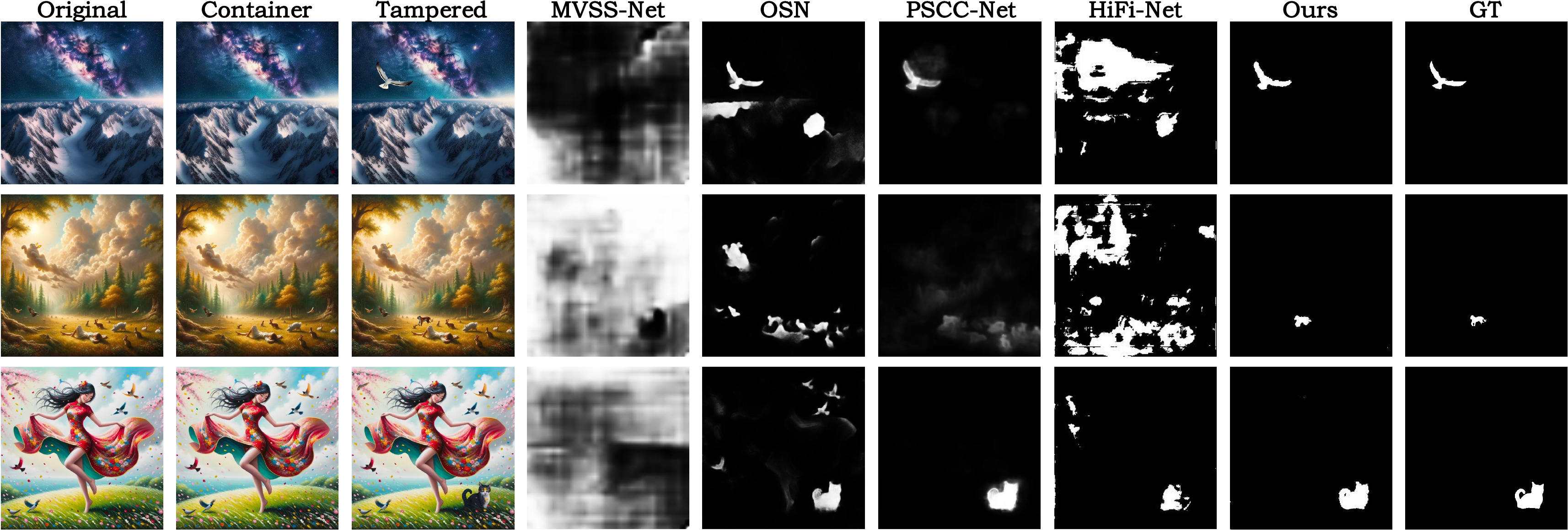}
	\caption{Localization results of our EditGuard and other methods~\cite{dong2022mvss, wu2022robust, liu2022pscc, guo2023hierarchical} on the constructed AGE-Set-F. The original images are generated from Dalle3~\cite{dalle3}. The tampered images are manipulated by Adobe Firefly. It can be seen that our method can also achieve great localization performance in AI-generated images and mature commercial AIGC filling software, which shows the practicality of our method. Zoom in for a better view.}
 \vspace{-15pt}
    \label{firefly}
\end{figure*}
\subsection{Why Choose an Asymmetric Network over a Flow-based Model for Bit Encryption and Recovery?}
In the framework of our EditGuard, we employed asymmetric U-shaped network BEM (Bit Encryption Module) and BRM (Bit Recovery Module) for modulating and decoding bit information. A more intuitive approach would be to use a flow model~\cite{fang2023flow} to hide and recover bits, similar to the image hiding and decoding modules. However, as shown in Tab.~\ref{robustness}, we found that although flow models achieve high localization accuracy and fidelity in the absence of degradation, their bit reconstruction accuracy and robustness are relatively unsatisfactory, just slightly above 95\%. Additionally, the performance of our variant significantly deteriorates when subjected to some degradation. Therefore, we ultimately opted for the more flexible approach of using an asymmetric encoder-decoder for copyright protection, which reflects the non-trivial nature of our network design.
\begin{table}[ht]
\centering
\caption{Localization and bit recovery performance of our EditGuard and the variant of our framework under different degradation.}
\vspace{-5pt}
\resizebox{\linewidth}{!}{
\begin{tabular}{c|c|c|cc|ccc|c}
\toprule[1.5pt]
\multirow{2}{*}{Methods} &\multirow{2}{*}{Metrics} & \multirow{2}{*}{Clean} & \multicolumn{2}{c|}{Gaussian Noise} & \multicolumn{3}{c|}{JPEG} & \multirow{2}{*}{Poisson} \\ \cline{4-8} 
                         & &                        & $\sigma$=1             & $\sigma$=5               & $Q$=70    & $Q$=80   & $Q$=90   & \\ \hline
\multirow{2}{*}{Ours-variant}     & F1 &0.984	&0.917	&0.913	&0.885 &0.908 &0.921	&0.908 \\
&BA(\%) &95.24 &95.15 &95.13 &94.63 &94.75    &95.10	&94.95 \\ \hline
\multirow{2}{*}{EditGuard (Ours)} & F1 & 0.966  & 0.937  & 0.932  & 0.920   & 0.920  & 0.925  & 0.943   \\
&BA(\%) & 99.95  & 99.94  & 99.37 & 97.69 & 98.16 & 98.23  & 99.91   \\ \bottomrule[1.5pt]
\end{tabular}}
\label{robustness}
\end{table}


\section{Limitations and Future Works}
\label{sec: limitations}
Our EditGuard provides a versatile watermarking solution to jointly achieve tamper localization and copyright protection. Simultaneously, we design a united image-bit steganography network (IBSN) tailored for proactive dual forensics. However, there are several limitations that await future research and improvement:
\begin{itemize}
    \item Currently, our selection of localization watermarks is still empirical. In the future, we hope to learn an optimal localization watermark via end-to-end optimization. Meanwhile, we expect to use a grayscale image, instead of a three-channel RGB image, as the localization watermark. This approach is intended to reduce the information capacity of our IBSN and further enhance its robustness.
    \item Our framework can be effectively extended to forensics and copyright protection of video~\cite{wu2023tune}, hyperspectral data~\cite{zhang2023progressive, zhang2022herosnet}, and 3D scenes~\cite{haque2023instruct}. The cross-modal verification of audio and temporal information has the potential to significantly enhance the localization accuracy of our method. At the same time, our framework will also expand upon the existing framework to include some additional functionalities, such as the self-recovery of original images~\cite{ying2023learning}.
\end{itemize}

\section{More Visual Results}
\begin{figure*}[t!]
	\centering
	\includegraphics[width=1\linewidth]{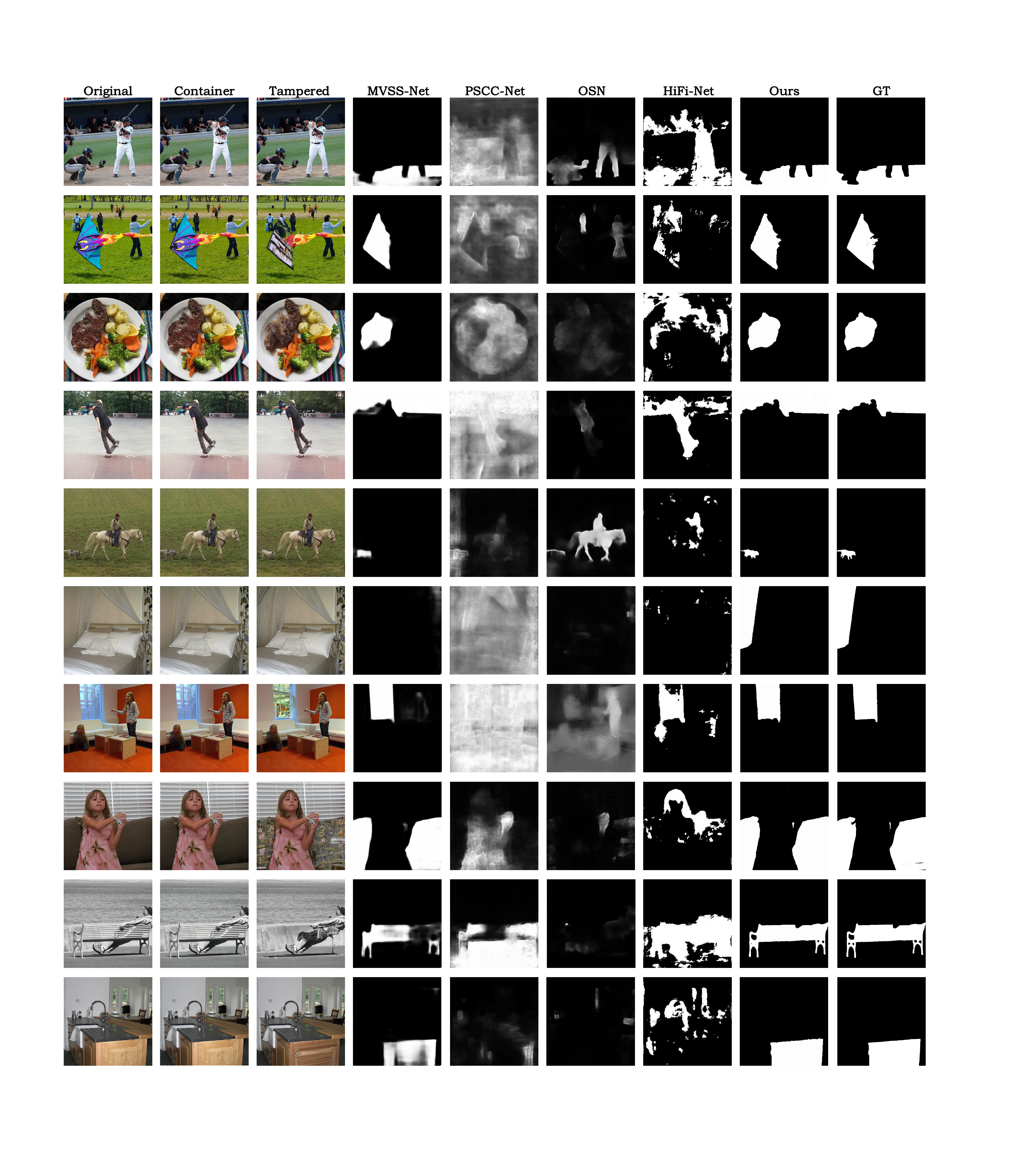}
	\caption{Visual comparisons of our EditGuard and other tamper localization methods~\cite{dong2022mvss, wu2022robust, liu2022pscc, guo2023hierarchical} on our constructed AGE-Set-C. Clearly, our method can pinpoint pixel-wise tampered areas, whereas other methods often struggle to effectively handle AIGC-based tampering. Zoom in for a better view.}
 \vspace{-25pt}
    \label{sd}
\end{figure*}

\begin{figure*}[t!]
	\centering
	\includegraphics[width=1\linewidth]{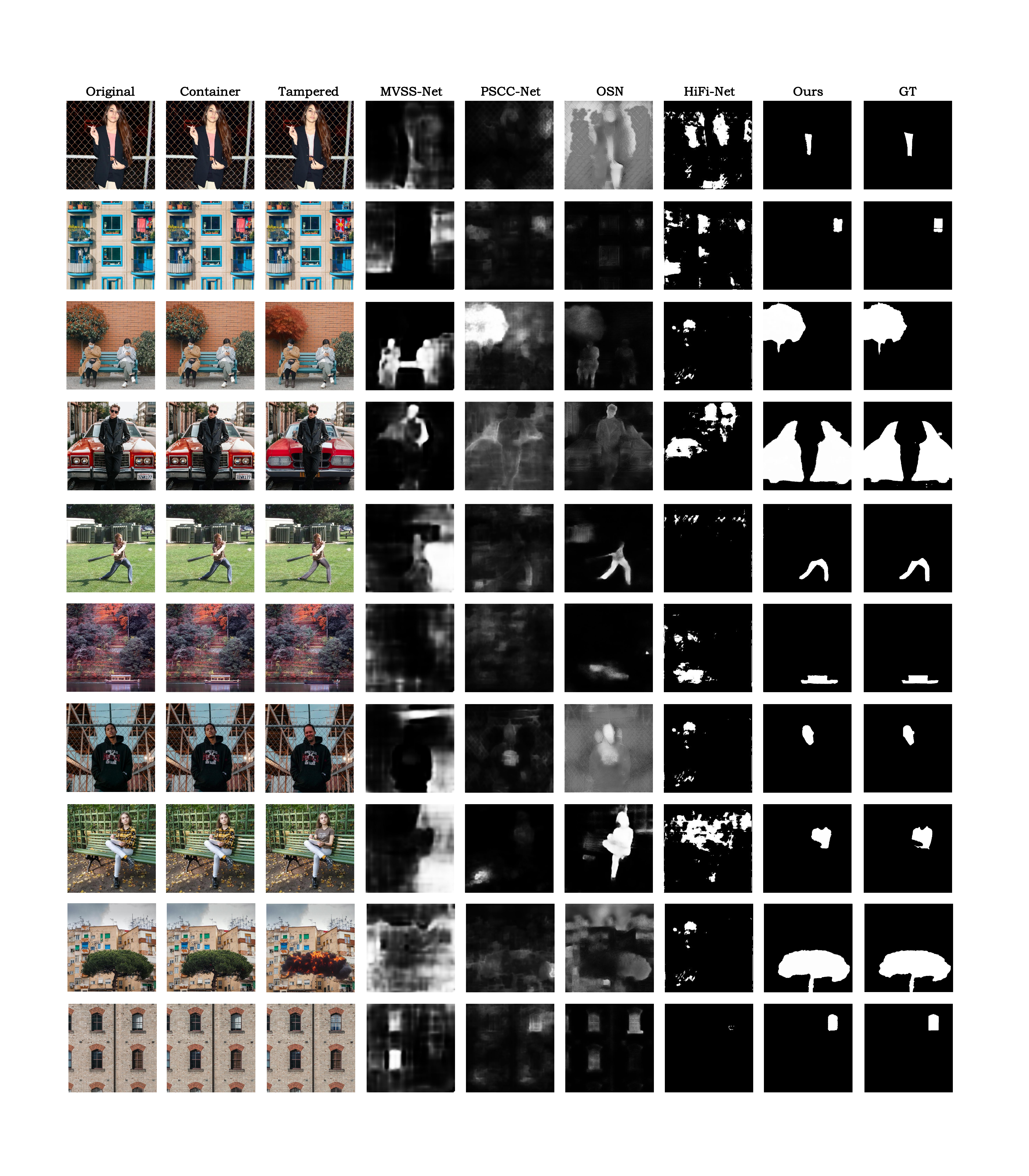}
	\caption{Localization results of our EditGuard and other methods~\cite{dong2022mvss, wu2022robust, liu2022pscc, guo2023hierarchical} on the constructed AGE-Set-F. The tampered images are manipulated by AIGC-based editing software with some customized models and well-designed prompts, and are hard for the naked eye to detect. In these cases, almost all methods fail but our EditGuard can still precisely produce the tampered masks. Zoom in for a better view.}
 \vspace{-15pt}
    \label{agef}
\end{figure*}

We present more visual results of our EditGuard and other competitive methods on our constructed AGE-Set-C (in Fig.~\ref{sd}) and AGE-Set-F (in Fig.~\ref{firefly} and Fig.~\ref{agef}). We found that our method can consistently achieve precise localization performance whether on batch-processed AGE-Set-C or careful editing with encapsulated editing software. However, almost none of the other methods can achieve comparable localization accuracy and generalization ability.

\label{sec: visual}

\newpage

\clearpage
{
    \small
    \bibliographystyle{ieeenat_fullname}
    \bibliography{main}
}


\end{document}